\documentclass[review,11pt]{elsarticle}
\usepackage{caption}
\usepackage[utf8]{inputenc}
\usepackage[english]{babel}
\usepackage{amssymb}
\usepackage{amsthm}
\usepackage{notoccite}
\usepackage{bm}
\usepackage{algorithmicx}
\usepackage{lscape}
\usepackage[noend]{algpseudocode}
\usepackage[flushleft]{threeparttable}
\usepackage{optidef}
\usepackage{booktabs}
\usepackage{lscape}
\DeclareUnicodeCharacter{2212}{-}
\usepackage[table,xcdraw]{xcolor}
\usepackage{rotating,tabularx}
\usepackage[nottoc]{tocbibind}

\usepackage{amssymb}
\usepackage{amsthm}
\usepackage{algorithmicx}
\usepackage{lscape}
\usepackage{adjustbox}
\usepackage[noend]{algpseudocode}
\usepackage[flushleft]{threeparttable}
\usepackage{booktabs}
\usepackage{amsthm}
\usepackage{algorithm}
\usepackage[noend]{algpseudocode}
\usepackage{lipsum,hyperref}
\usepackage{mathbbol}
\usepackage{subfigure}
\hypersetup{
    colorlinks=true,
    linkcolor=cyan,
    filecolor=magenta,      
    urlcolor=cyan,
}

\newtheorem{proposition}{Proposition}

\newproof{pf}{Proof}
\newcommand{\red}{\textcolor{black}}
\renewcommand\vec{\mathbf}

\usepackage{lineno,hyperref}
\modulolinenumbers[5]

\begin{document}

\begin{frontmatter}

\title{On multivariate randomized classification trees: $l_0$-based sparsity, VC~dimension and decomposition methods}

\author[mymainaddress]{Edoardo Amaldi}
\ead{edoardo.amaldi@polimi.it}
\author[mymainaddress]{Antonio Consolo\corref{mycorrespondingauthor}}
\ead{antonio.consolo@polimi.it}
\author[mysecondaryaddress]{Andrea Manno}
\ead{andrea.manno@polimi.it}


\cortext[mycorrespondingauthor]{Corresponding author}

\address[mymainaddress]{DEIB, Politecnico di Milano, Milano, Italy}
\address[mysecondaryaddress]{Centro di Eccellenza DEWS, DISIM, Università degli Studi dell’Aquila, L'Aquila, Italy}

\begin{abstract}

Decision trees are widely-used classification and regression models because of their interpretability and good accuracy. 
Classical methods such as CART are based on greedy approaches but a growing attention has recently been devoted to optimal decision trees.
We investigate the nonlinear continuous optimization formulation proposed in Blanquero et al.  (EJOR, vol. 284, 2020; COR, vol. 132, 2021) for (sparse) optimal randomized classification trees.
Sparsity is important not only for feature selection but also to improve interpretability. We first consider alternative methods to sparsify such trees based on concave approximations of the $l_{0}$ ``norm". Promising results are obtained on 24 datasets in comparison with $l_1$ and $l_{\infty}$ regularizations. Then, we derive bounds on the VC dimension of multivariate randomized classification trees. Finally, since training is computationally challenging for large datasets, we propose a  general decomposition scheme and an efficient version of it. 
Experiments on larger datasets show that the proposed decomposition method is able to significantly reduce the training times without compromising the accuracy. 
\end{abstract}

\begin{keyword}
Machine Learning \sep  randomized classification trees \sep sparsity \sep decomposition methods \sep  nonlinear programming 
\end{keyword}

\end{frontmatter}


\section{Introduction}

Decision trees are 
popular classification and regression models in the areas of Machine Learning (ML) and Data Mining. Because of their interpretability and their good accuracy, they are applied in a number of fields ranging from Medicine \textcolor{black}{(see e.g. \cite{podgorelec2002decision,tsien2000multiple,Chelazzi2021})} to Business Analytics \textcolor{black}{(see e.g. \cite{ghatasheh2014business,ghodselahi2011application,ouahilal2016comparative})}. 


Since building optimal binary decision trees is NP-hard \cite{laurent1976constructing} and large-scale datasets are often of interest, CART \cite{breiman2017classification} pioneering work and later extensions, such as ID3 \cite{quinlan1986induction} and C4.5\cite{quinlan2014c4}, adopt a greedy and top-down approach aiming at minimizing an impurity measure.
Then a pruning phase is used to simplify the tree topology in order to reduce overfitting and to obtain a more interpretable model. 


Due to the 
remarkable progress in the computational performance of Mixed-Integer Linear Optimization (MILO) and nonlinear optimization solvers, decision trees have been revisited during the last decade. 


Most previous work on optimal classification trees is concerned with deterministic trees where each input vector is univocally associated to a single class. In \cite{bennett1996optimal} an extreme point tabu search method is described to minimize the misclassification error of all decisions in a given tree concurrently.
In \cite{bertsimas2017optimal,dunn2018optimal}, a MILO formulation and a local search approach are proposed for constructing optimal multivariate classification trees. 
In \cite{gunluk2021optimal} an integer programming formulation is presented to design binary classification trees for categorical data. 
An efficient integer optimization encoding is proposed in \cite{verwer2019learning} to construct classification and regression trees of depth $D$ with univariate decisions at the branch nodes. In \cite{firat2018column} a column generation heuristic 
is described to build univariate binary classification trees for larger datasets. 
A dynamic programming and search algorithm is presented in \cite{demirovic2020murtree} for constructing optimal univariate decision trees. \textcolor{black}{In \cite{aghaei2020learning} the authors propose a flow-based MILO formulation for optimal \textcolor{black}{univariate} classification trees where they exploit the problem structure and max-flow/min-cut duality to derive a Benders’ decomposition method for handling large datasets.}

Recently, in \cite{blanquero2018optimal,blanquero2020sparsity}, a novel continuous nonlinear optimization approach has been proposed to build (sparse) optimal multivariate randomized classification trees. 
At each branch node a random variable is generated to determine to which branch (left or right) an 
input vector is 
forwarded to. An appealing feature of multivariate randomized classification trees with respect to deterministic ones is their probabilistic nature in terms of the posterior probability. 
Since such trees involve only continuous variables, they can be trained with a continuous constrained nonlinear programming solver. Although the formulation is nonconvex, some available solvers are guaranteed to converge to feasible solutions satisfying local optimality conditions.
In \cite{blanquero2020sparsity}, sparsity of multivariate randomized classification trees is achieved by adding $l_{1}$ and $l_{\infty}$ regularization terms to the objective function. The interested reader is referred to \cite{carrizosa2021mathematical} for a survey on 
optimal decision trees.

In this work, we investigate sparse multivariate randomized classification trees. In particular, we describe alternative sparsification strategies based on concave approximations of the $l_0$ ``norm" \footnote{$l_0$ is not a proper norm since it does not satisfy the absolute homogeneity assumption.} and we evaluate on 24 datasets their potential benefits compared with the above-mentioned regularizations. 
Then we discuss a theoretical aspect of 
such trees, namely their Vapnik-Chervonenkis (VC) dimension \cite{blumer1989learnability}.
Finally, we propose a general proximal point decomposition scheme to reduce the training times 
for larger datasets. After commenting on the asymptotic convergence, we present an efficient specific version 
of the decomposition scheme and test it on 5 datasets in comparison with the original, not decomposed version. 

The remainder of the paper is organized as follows. In Section \ref{sec:ORCTs} we briefly summarize the formulation proposed in \cite{blanquero2018optimal} for optimal randomized classification trees.
In Section \ref{sec:sparsity}, after recalling how sparsity is pursued in \cite{blanquero2020sparsity}, we present alternative $l_0$-based strategies. 
The computational results are reported and discussed in Section \ref{sec:comp_results}.
Section \ref{sec:VC-dimension} is devoted to upper and lower bounds on the VC dimension of multivariate randomized classification trees.
In Section \ref{sec:decopmosition}, the general decomposition scheme is described and the results obtained with a specific version are reported. Finally, Section \ref{sec:concluding} contains some concluding remarks.

\section{Optimal randomized classification trees}
\label{sec:ORCTs}


We briefly recall the nonlinear continuous optimization formulation proposed in \cite{blanquero2018optimal} to train Optimal Randomized Classification Trees (ORCTs). 


Consider a training set $I=\left \{ (\mathbf{x}_{i} ,y_{i})\right \}_{1\leq i \leq N}$ consisting of $N$ samples, where $\mathbf{x}_{i}\in {\mathbb R}^p$ is the $p$-dimensional vector of predictor variables and $y_{i}\in \left \{ 1,\ldots,K \right \}$ the associated class label. 



\begin{figure}[h]
\centering
\includegraphics[scale=0.5]{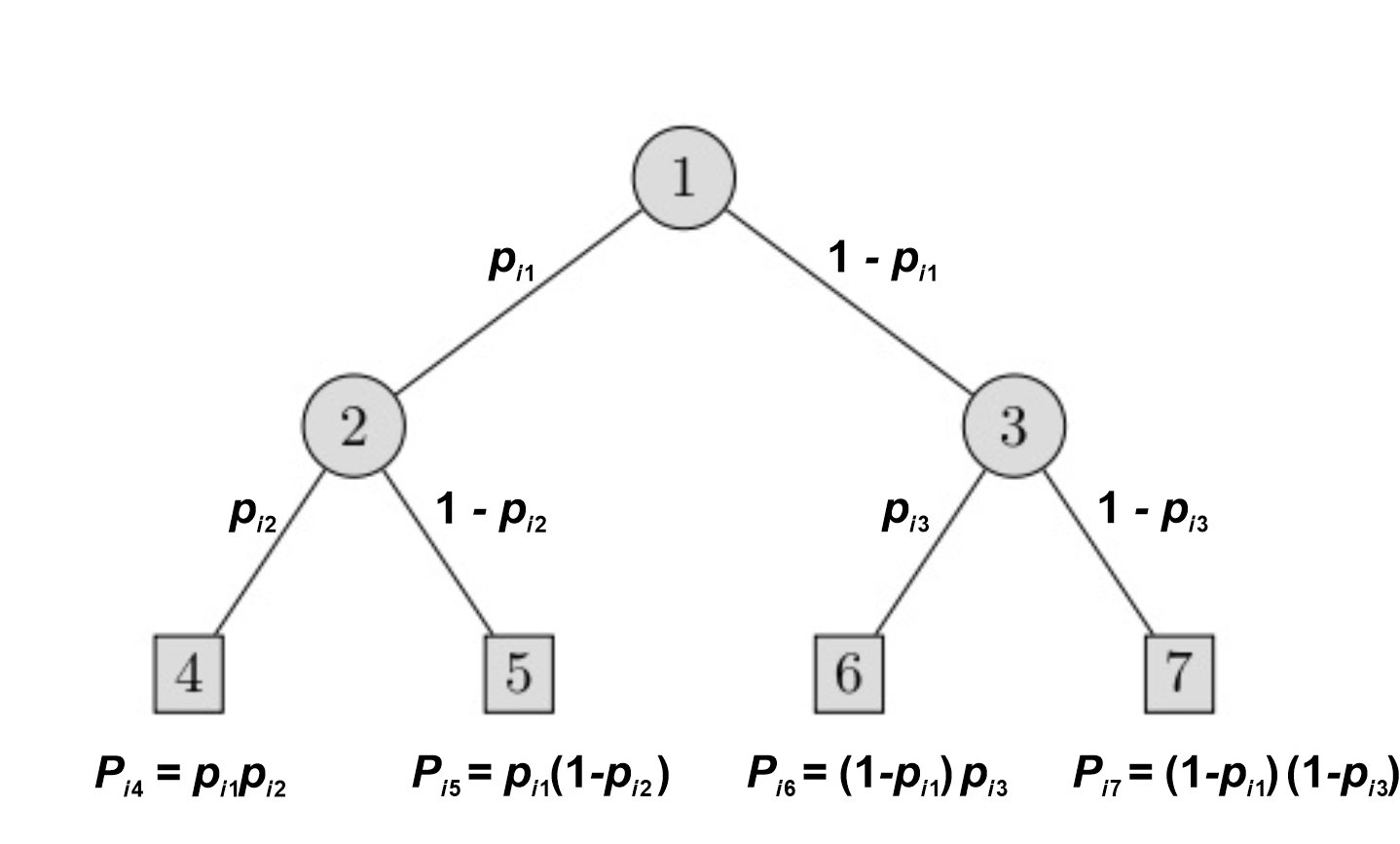}
\caption{\label{fig:fig12}An example of an Randomized Classification Tree with depth $D=2$.}
\end{figure}

Randomized Classification Trees are maximal binary trees of a given depth $D$, with $D \geq 1$.
Let $\tau_{L}$ and $\tau_{B}$ denote, respectively, the set of leaf nodes and of branch nodes.
%
At branch nodes multivariate 
(hyperplane) splits are performed according to a probabilistic splitting rule.
The probability of taking a branch is determined by a univariate cumulative density function (CDF) evaluated over a linear combination of the predictor variables. 
%
%
More precisely, for each input vector $\mathbf{x}_{i}$ with $i \in \{1,....,N\}$ and each branch node $t \in \tau_{B}$ the probability of taking the left branch is given by

$$ p_{it}=  F_{\gamma} 
(\frac{1}{p}\sum_{j=1}^{p}a_{jt}x_{ij}-\mu_{t}), $$
where the coefficients $a_{jt}\in \left [-1,1 \right ]$ and $\mu _{t}\in \left [ -1,1 \right ]$ are the decision variables, and the logistic CDF
$$ F_{\gamma} 
(v)=\frac{1}{1+e^{\left ( - \gamma v  \right )}} $$
with parameter $\gamma>0$ is considered.
The right branch is taken with probability $1-p_{it}$. See Figure \ref{fig:fig12} for an example of a Randomized Classification Tree of depth $D=2$.

Since the logistic CDF induces a soft splitting rule 
at each branch node, all 
input vectors in the training set fall into every leaf node with a certain probability. 
%
%
The probability that an input vector $\mathbf{x}_{i}$ with $i \in \{1,....,N\}$ falls into leaf node $t \in \tau_{L}$ is then given by
\begin{align}
\label{eq:prob-constr}
\qquad
P_{it} = 
\prod_{t_{l}\in N_{L(t)}}p_{it_{l}}\;\prod_{t_{r}\in N_{R(t)}}(1-p_{it_{r}}),
\end{align}
where $N_{L}(t)$ denotes the set of ancestor nodes of leaf node $t$ whose left branch belongs to the path from the root to $t$, while $N_{R}(t)$ the set of ancestor nodes for the right branch.







For each leaf node $t \in \tau_{L}$ and class label $k \in \{1,....,K\}$, let the binary decision variable $c_{kt}$ be equal to $1$ if at node $t$ all input vectors are assigned to the class label $k$, and $0$ otherwise. 

For each sample $(\mathbf{x}_{i},y_{i})$ with $i \in \{1,....,N\}$ and class label $k \in \{1,....K\}$, let the parameter $w_{y_{i} k} \geq 0$ denote the misclassification cost when classifying $\mathbf{x}_{i}$ in class $k$.

Then the problem of minimizing the expected misclassification error of the randomized classification tree over the training set can be formulated as the following mixed-integer nonlinear optimization problem:
\vspace{-20pt}
\begin{mini!}|s|[2]                   
    {}                               
    {  \sum_{i=1}^{N}\sum_{t\in \tau_{L}}P_{it}
\sum_{k=1}^{K}w_{y_{i}k}c_{kt}  \label{eq:obj-fct-orct}} 
    {}             
    {}                                
    \addConstraint{\sum_{k=1}^{K}c_{kt}}{=1 \qquad t \in \tau_{L} \label{eq:assign-leaf-orct}}    
    \addConstraint{\sum_{t \in \tau_{L}}c_{kt}}{\geq  1 \quad   k \in \{1,....,K\} \label{eq:assign-class-orct}}  
    \addConstraint{a_{jt}\in [-1,1],\;\;\mu_{t}}{\in [-1,1] \quad  j \in \{1,\ldots,p\}, t \in \tau_{B}  \label{eq:var1-orct}}  
    \addConstraint{c_{kt}}{\in \left \{ 0,1 \right \} \quad k \in \{1,\ldots,K\}, t \in \tau_{L},  \label{eq:var3-orct}}
\end{mini!}
where constraints \eqref{eq:assign-leaf-orct} ensure that each leaf node is assigned to exactly one class label, and constraints \eqref{eq:assign-class-orct} that each class label $k$ is associated with at least one leaf node. Remember that, for every pair $i \in \{1,\ldots,N\}$ and $t \in \tau_{L}$, the probability $P_{it}$ is a non linear function function of the variables $a_{jt'}$ and $\mu_{t'}$ with $j \in \{1,\ldots,p\}$ and $t' \in N_{L}(t) \cup N_{R}(t)$.

As shown in \cite{blanquero2018optimal}, the integrality of the binary variables $c_{kt}$ can be relaxed because the resulting nonlinear continuous formulation admits an optimal integer solution. 
Thus \eqref{eq:var3-orct} is substituted with 
\begin{align}
\label{eq:var3-orct-continuous} 
c_{kt} \in \left [ 0,1 \right ] \quad k \in \{1,\ldots,K\}, t \in \tau_{L}, 
\end{align}
where the variable $c_{kt}$ can be viewed as the probability that a leaf node $t$ is assigned to class label $k$.


After the training phase, that is, after solving the above nonlinear optimization formulation \eqref{eq:obj-fct-orct}-\eqref{eq:var3-orct}, the class for a new unlabeled input vector $\mathbf{x} \in {\mathbb R}^{p}$ is predicted by 
assigning $\mathbf{x}$ to the class for which the estimate $\sum_{t \in \tau_{L}} c_{kt} P_{xt}$ of the probability that $x$ belongs to class label $k$ is maximum.

As shown in \cite{blanquero2018optimal}, the above formulation can be easily amended to account for other constraints such as minimum correct classification rates for different classes.




\section{Sparse optimal randomized classification trees}
\label{sec:sparsity}


In many practical classification tasks the input vectors $\mathbf{x}_{i} \in {\mathbb R}^{p}$ include a large number $p$ of predictor variables, referred to as \emph{features}.  
The degree of sparsity of a model depends on both the number of features that are actually used and the number of nonzero parameters.
Sparse models are important not only because they identify a subset of most relevant features ({\it feature selection}) but also because they are more interpretable. Interpretability is a crucial issue for ML methods since it broadens their range of applicability. Moreover, according to Occam's razor principle, simpler models also tend to avoid overfitting and to yield a smaller generalization error, i.e., a higher accuracy on input vectors not included in the training set. 

In the Statistics and ML literature several approaches have been proposed to seek parsimonious models. A popular one consists in adding to the objective function ad hoc regularization terms inducing sparsity. For instance, in Lasso\footnote{Lasso stands for least absolute shrinkage and selection operator.} 
regression (see e.g. \cite{tibshirani1996regression}) penalizing the $l_1$ norm of the parameters vector allows to perform both feature selection and regularization, which in turn enhance both prediction accuracy and interpretability.  

In the context of classification trees, the degree of sparsity is related to number of features actually involved in the splitting rules implemented at the branch nodes. Two different types of sparsity naturally arise. Local sparsity corresponds to the total number of features occurring in the 
hyperplane splits at the branch nodes, while global sparsity corresponds to the number of features occurring across the whole tree. 


In \cite{blanquero2020sparsity} the authors promote the sparsity of optimal randomized classification trees 
by adding to the expected misclassification error over the training set two regularization terms based on polyhedral norms of the parameters vector. 
Adopting the $l_{1}$ norm for local sparsity and the $l_{\infty}$ norm for global sparsity, the overall objective function in sparse ORCT is:
\begin{equation}
\label{eq:sparse-orct-blanquero}
\sum_{i=1}^{N}\sum_{t \in \tau_{L}}P_{it}\sum_{k=1}^{K}w_{y_{i}k}c_{kt} 
+\lambda^{L}\sum_{j=1}^{p}\left \| \mathbf{a}_{j\cdot} \right \|_{1}+\lambda^{G}\sum_{j=1}^{p}\left \| \mathbf{a}_{j\cdot} \right \|_{\infty}
\end{equation}
where $\mathbf{a}_{j.}$ denotes the $|\tau_{B}|$-dimensional vector of the coefficients of the $j$-th feature for all branch nodes $t \in \tau_{B}$. 
An equivalent smooth formulation can be easily obtained by rewriting the two regularization terms using additional variables and constraints. 

From now on we will use the acronym MRCTs for Multivariate Randomized Classification Trees with possibly other objective functions.









\subsection{Sparse multivariate randomized classification trees via approximate $l_0$ regularization}
\label{subsec:l_0-sparsity}

\medskip
To induce local and global sparsity in MRCTs, we consider penalizing the $l_0$ ``norm" of the parameters vector rather than the $l_{1}$ and $l_{\infty}$ norms. The $l_{0}$ ``norm" of 
a vector $\mathbf{v} \in \mathbb{R}^{n}$ is 
the number of nonzero components of $\mathbf{v}$, namely    
$$ \left \| \mathbf{v} \right \|_{0}=\sum_{l=1}^{n} \mathbb{ 1}_{\mathbb{R}^{+}}(|v_l|), $$
where $\mathbb{ 1}_{\mathbb{R}^{+}}(u)$ denotes the step function with
$\mathbb{ 1}_{\mathbb{R}^{+}}(u)=0$ for $u \leq 0$ and $\mathbb{ 1}_{\mathbb{R}^{+}}(u)=1$ for $u>0$.
For local sparsity, we add to the loss function \eqref{eq:obj-fct-orct} the regularization term
$$ \sum\limits_{j=1}^{p}\left \| \mathbf{a}_{j.} \right \|_{0}=\sum\limits_{j=1}^{p}\sum_{t \in \tau_{B}} \mathbb{ 1}_{\mathbb{R}^{+}}\left ( \left | a_{jt} \right | \right ). $$ 
which counts the total number of predictor variables (features) actually involved in the 
multivariate (hyperplane) splits implemented at the branch nodes.
For global sparsity, we also add to the loss function \eqref{eq:obj-fct-orct} the regularization term:
$$ \left \| \bm{\beta}  \right \|_{0}=\sum_{j=1}^{p} \mathbb{ 1}_{\mathbb{R}^{+}}\left (  \beta_{j}   \right ) $$
where the new variables $\beta_{j} \in \left[0,1\right]$ are subject to 
\begin{align}
-\beta_{j} \leq a_{jt} \leq \beta_{j} \;\;  \quad &  j \in \{1,\ldots,p\},\ t \in \tau_{B}. \label{eq:constraints-beta-1} 
\end{align} 

\noindent This second term is equal to the number of features 
that are actually used across the whole tree. 

Although $l_0$ regularization is a natural way to induce local and global sparsity, the resulting nonlinear optimization problem is more challenging than the ones involving the $l_1$ and $l_\infty$ norms because the step function $\mathbb{ 1}_{\mathbb{R}^{+}}(u)$ is discontinuous. 
Since the $l_0$ penalty terms make the overall objective function non-smooth, we consider continuously differentiable concave approximations.
This approach was introduced in \cite{bradley1998feature} for linear classification models and further developed in \cite{weston2003use}
and in \cite{rinaldi2010feature}. 
However, unlike in these and other previous works, the training of sparse MRCTs cannot be reduced to an overall concave optimization problem.

Similarly to \cite{bradley1998feature}, we replace the discontinuous step function $\mathbb{ 1}_{\mathbb{R}^{+}}(u)$ with the smooth concave exponential approximation $1-e^{-\alpha u}$ on the non-negative real line $u \geq 0$, with parameter $\alpha > 0$. This leads to the following approximate $l_0$ regularization term for local sparsity
\begin{equation}
\label{eq:l_0-approx-local}
\sum_{j=1}^{p}\sum_{t\in \tau _{B}}(1-e^{-\alpha z_{jt}}),
\end{equation} 
where the additional variables $z_{jt} \in \left[0,1\right]$ satisfy constraints 
\begin{align}
-z_{jt} \leq a_{jt} \leq z_{jt}  \quad \quad & j \in \{1,\dots,p\},\ t \in \tau_{B}, \label{eq:constraints-zjt-1}
\end{align} 
and to the following approximate $l_0$ regularization term for global sparsity
\begin{equation}
\label{eq:l_0-approx-global}
\sum_{j=1}^{p}(1-e^{-\alpha \beta_{j}}),
\end{equation} 
where the additional variables $\beta_{j} \in \left[0,1\right]$ satisfy constraints (\ref{eq:constraints-beta-1}).\par
Thus we obtain the alternative formulation for sparse multivariate randomized classification trees:  
\begin{equation}
\begin{matrix}
    & \min  \sum\limits_{i=1}^{N}\sum\limits_{t \in \tau_{L}}P_{it}\sum\limits_{k=1}^{K}w_{y_{i}k}c_{kt}+\lambda_{0}^{L}\sum\limits_{j=1}^{p}\sum\limits_{t\in \tau_{B}}(1-e^{-\alpha z_{jt}})+\lambda_{0}^{G}\sum\limits_{j=1}^{p}(1-e^{-\alpha \beta_{j}})  \\
    & \sum\limits_{k=1}^{K}c_{kt}=1 \qquad t \in \tau_{L} \\    
    & \sum\limits_{t \in \tau_{L}}c_{kt}\geq  1 \quad   k \in \{1,....,K\} \\ \label{eq:l_0-base-formulation}
    & -\beta_{j} \leq a_{jt} \leq \beta_{j}, \;\;\beta_{j} \in \left[ 0,1 \right]  \quad  j \in \{1,\ldots,p\},\ t \in \tau_{B} \\
    & -z_{jt} \leq a_{jt} \leq z_{jt}, \quad z_{jt} \in \left[ 0,1 \right]  \quad \quad j \in \{1,\dots,p\},\ t \in \tau_{B} \\
    & a_{jt}\in [-1,1],\;\;\mu_{t}\in [-1,1] \quad  j \in \{1,\ldots,p\}, t \in \tau_{B}   \\ 
    & c_{kt}\in \left[ 0,1 \right] \quad k \in \{1,\ldots,K\}, t \in \tau_{L},
\end{matrix}
\end{equation}
where $\lambda_{0}^{L} \geq0$ and $\lambda_{0}^{G} \geq0$ are, respectively, the local and global sparsity regularization parameters, and the additional variables $\beta_{j},z_{jt} \in \left[0,1\right]$ satisfy the 
corresponding constraints (\ref{eq:constraints-beta-1}) and (\ref{eq:constraints-zjt-1}).

In \cite{blanquero2020sparsity} the authors establish for the sparse ORCT formulation lower bounds on the regularization parameters $\lambda^L$ and $\lambda^G$ to ensure that the most sparse tree possible (i.e. $\textbf{a}^*=\textbf{0}$) is a stationary point. Here we derive in a different way a similar result for $\lambda_{0}^{L}$ and $\lambda_{0}^{G}$ of formulation \eqref{eq:l_0-base-formulation}. 

\begin{proposition}
	\label{prop:regularization}
	Assume that $\lambda_{0}^{L}$ and $\lambda_{0}^{G}$ are such that
	$$\lambda_{0}^{L} + \lambda_{0}^{G} \geq \max_{j=1,...,p,t\in \tau_B} \frac{|\xi_{jt}(0)|}{\alpha},$$
	where $\xi_{jt}(0)$ represents the partial derivative of the objective function \eqref{eq:obj-fct-orct} with respect to $a_{jt}$ evaluated at  zero. Then a stationary point ($\textbf{a}^*,\;\bm{\mu^{*}},\textbf{c}^*$) for 
	formulation \eqref{eq:l_0-base-formulation} exists with $\textbf{a}^*=\textbf{0}$. 	
\end{proposition}
\begin{proof}
From Theorem 1 in \cite{blanquero2018optimal} we know that 
formulation \eqref{eq:l_0-base-formulation} admits an optimal solution $(\mathbf{a}^*,\bm{\mu^*},\mathbf{c}^*)$ such that $c^{*}_{kt} \in \{0,1\} \; \forall k = 1,...,K, t \in \tau_L$. 
For the sake of proof simplicity, we rewrite the objective function in \eqref{eq:l_0-base-formulation}
without the $z_{jt}$ variables: 
\begin{equation}\label{eq:l0_absolute_objective}
\sum\limits_{i=1}^{N}\sum\limits_{t \in \tau_{L}}P_{it}\sum\limits_{k=1}^{K}w_{y_{i}k}c_{kt}+\lambda_{0}^{L}\sum\limits_{j=1}^{p}\sum\limits_{t\in \tau_{B}}(1-e^{-\alpha |a_{jt}|})+\lambda_{0}^{G}\sum\limits_{j=1}^{p}(1-e^{-\alpha \beta_{j}})
\end{equation}  
and we consider the formulation where \eqref{eq:l0_absolute_objective} is minimized subject to all the constraints in formulation \eqref{eq:l_0-base-formulation} except constraints \eqref{eq:constraints-zjt-1} which involve the $z_{jt}$ variables. The two formulations are clearly equivalent since at optimality for each pair of inequality constraints (\ref{eq:constraints-zjt-1}) one is satisfied with equality.

We distinguish three cases: (i) $\lambda_{0}^{L}>0$ and $\lambda_{0}^{G}=0$, (ii) $\lambda_{0}^{L}=0$ and $\lambda_{0}^{G}>0$, and (iii) $\lambda_{0}^{L}>0$ and $\lambda_{0}^{G}>0$. First, we prove the result for case (i), then we show how it can be easily extended to cases (ii) and (iii).

Let us consider the solution 
$(\mathbf{0}^*,\bm{\mu^*},\mathbf{c}^*)$.
By assuming $ \Xi = \sum_{i=1}^{N}\sum_{t \in \tau_{L}}P_{it}\sum_{k=1}^{K}w_{y^{i}k}c_{kt}$ (the first term of \eqref{eq:l0_absolute_objective}), for a fixed predictor variable index $\bar{j}$ and branching node index $\bar{t}$, the partial derivative of $\Xi$ with respect to $a_{\bar{j}\bar{t}}$ (denoted as $\xi_{\bar{j}\bar{t}}(a_{\bar{j}\bar{t}})$) is:
\begin{equation*}\begin{split}
    {
    \frac{\partial \Xi}{\partial a_{\bar{j}\bar{t}}} = \xi_{\bar{j}\bar{t}}(a_{\bar{j}\bar{t}}) =  \sum_{i=1}^{N}\sum_{t \in \tau_{L}(\bar{t})} \ \prod_{l \in N_L(t):l \neq \bar{t}} p_{il}\prod_{r \in N_R(t):r \neq \bar{t}} (1-p_{ir})(-1)^{b^{t\bar{t}}} \sum_{k=1}^{K}w_{y^{i}k}c_{kt}
    \frac{\gamma e^{-\gamma(\frac{1}{p}\sum_{j=1}^{p}a_{j\bar{t}}x_{ij}-\mu_{\bar{t}})}}{p(1+e^{-\gamma(\frac{1}{p}\sum_{j=1}^{p}a_{j\bar{t}}x_{ij}-\mu_{\bar{t}})})^2} x_{i\bar{j}}
   }
\end{split}
\end{equation*}


\noindent where $\tau_{L}(\bar{t})$ is the set of all the leaf nodes descendant from node $\bar{t}$,  
formally $$\tau_{L}(\bar{t}) = \{ t : t \in \tau_L , \bar{t} \in N_L(t) \lor \bar{t} \in N_R(t) \},$$ and \begin{math}
  b^{t\bar{t}}=\left\{
    \begin{array}{ll}
      1 & \mbox{if $\bar{t} \in N_R(t)$},\\
      0 & \mbox{if $\bar{t} \in N_L(t)$}.
    \end{array}
  \right.
\end{math}
\medskip

As to the sparsity regularization terms, since by assumption $\lambda_{0}^{G}=0$, we only need to consider the local sparsity term $R^{loc} =\lambda_{0}^{L} \sum_{j=1}^{p}\sum_{t\in \tau _{B}}(1-e^{-\alpha | a_{jt}|})$ whose single component $R^{loc}_{\bar j \bar t}$ with respect to 
$a_{\bar{j}\bar{t}}$ can be rewritten as:
\begin{equation}\label{eq:Rloc}
    R^{loc}_{\bar j\bar t} =\lambda_{0}^{L}( 1 - e^{-\alpha |a_{\bar j \bar t}|} )= \lambda_{0}^{L}(-\sum_{k=1}^{\infty} \frac{(-\alpha |a_{\bar j \bar t}|)^k}{k!} )=  \lambda_{0}^{L}(\alpha |a_{\bar j \bar t}| + o(|a_{\bar j \bar t}|)).
\end{equation}
For $a_{\bar{j}\bar{t}} = 0$ the optimality condition is $0 \in \xi_{\bar{j}\bar{t}}(0) + \lambda_{0}^{L} \partial R^{loc}_{\bar{j}\bar{t}} $, where $\partial R^{loc}_{\bar{j}\bar{t}}$ is the subdifferential of $R^{loc}_{\bar j\bar t}$. 
$\partial R^{loc}_{\bar{j}\bar{t}}$ is equal to $[l(\alpha,a_{\bar{j}\bar{t}}),u(\alpha,a_{\bar{j}\bar{t}})]$, where $l(\alpha,a_{\bar{j}\bar{t}}) $ and $u(\alpha,a_{\bar{j}\bar{t}})$ are respectively the lower and upper bound of the subdifferential interval as functions of the parameter $\alpha$ and the 
variables
$a_{\bar j\bar t}$. When $a_{\bar{j}\bar{t}} = 0$, $\partial R^{loc}_{\bar{j}\bar{t}}$ coincides with the subdifferential of function $ \lambda_{0}^{L}\alpha |a_{\bar{j}\bar{t}}|$, therefore $\partial R^{loc}_{\bar{j}\bar{t}} = [-\lambda_{0}^{L}\alpha,\lambda_{0}^{L}\alpha]$. Thus, the optimality condition is $0 \in \left[\xi_{\bar{j}\bar{t}}(0) - \lambda_{0}^{L}\alpha,\xi_{\bar{j}\bar{t}}(0) + \lambda_{0}^{L}\alpha\right]$, hence 
\begin{equation}\label{eq:lb}
    \lambda_{0}^{L} \geq \frac{|\xi_{\bar{j}\bar{t}}(0)|}{\alpha}. 
\end{equation} By applying \eqref{eq:lb} to every possible pair $\bar j$ and $\bar t$ the result follows.
\par

Concerning case (ii) ($\lambda_{0}^{L}=0$), the only sparsity regularization term is $R^{glob} = \lambda_{0}^{G}\sum_{j=1}^{p}(1-e^{-\alpha \beta_j})$, where 
\vspace{-0.32cm}
\begin{align*}
\beta_j = \max\limits_{t \in \tau_B} |a_{jt}|,  \quad &  j \in \{1,\ldots,p\}. \end{align*} 
For every 
predictor variable index $\bar j$, we have $\beta_{\bar j}=|a_{\bar j \bar t(\bar j)}|$, where $t(\bar j)=\text{argmax}_{t \in \tau_B} |a_{\bar jt}|$. 
Thus the single component of the sparsification term 
corresponding to $\bar j$ amounts to 
\begin{equation}\label{eq:Rglob}
    R^{glob}_{\bar j} =\lambda_{0}^{G}( 1 - e^{-\alpha |a_{\bar j t(\bar j)}|}) = \lambda_{0}^{G}(-\sum_{k=1}^{\infty} \frac{(-\alpha |a_{\bar jt(\bar j)}|)^k}{k!}) = \lambda_{0}^{G}( \alpha |a_{\bar jt(\bar j)}| + o(|a_{\bar jt(\bar j)}|)).
\end{equation}
Since $a_{\bar jt}=0$ for every $t \in \tau_B$ at $(\mathbf{0}^*,\bm{\mu^*},\mathbf{c}^*)$, we can replace $t(\bar j)$ with any $t$ and we obtain the result by applying the same reasoning of case (i) (replacing $\lambda_{0}^{L}$ with $\lambda_{0}^{G}$). 

In case (iii), the sparsity regularization term is $R^{tot} = \lambda_{0}^{L}R^{loc}+ \lambda_{0}^{G} R^{glob}$, with for every pair $\bar j$ and $\bar t$ a component equal to
\begin{equation}\label{eq:Rtot}
\begin{matrix}
        R^{tot}_{\bar j\bar t} =\lambda_{0}^{L}( 1 - e^{-\alpha |a_{\bar j \bar t}|} )+ \lambda_{0}^{G}( 1 - e^{-\alpha |a_{\bar j \bar t}|})=  (\lambda_{0}^{L}+ \lambda_{0}^{G})( 1 - e^{-\alpha |a_{\bar j \bar t}|} ) = \\
         (\lambda_{0}^{L}+ \lambda_{0}^{G})(-\sum_{k=1}^{\infty} \frac{(-\alpha |a_{\bar j\bar t}|)^k}{k!})= (\lambda_{0}^{L}+ \lambda_{0}^{G})(\alpha |a_{\bar j \bar t}| + o(|a_{\bar j \bar t}|)).
\end{matrix}
\end{equation}
Notice that, in general, the global term $ \lambda_{0}^{G}( 1 - e^{-\alpha |a_{\bar j \bar t}|})$ of \eqref{eq:Rtot} is present only if $\bar t\equiv t(\bar j)$. However, as already pointed out, since $a_{\bar jt}=0$ for every $t \in \tau_B$,  \eqref{eq:Rtot} applies to any pair $\bar j$ and $\bar t$. Then, 
the result is obtained by applying the same reasoning of case (i), replacing $\lambda_{0}^{L}$ with $\lambda_{0}^{L}+\lambda_{0}^{G}$. 

%

\end{proof}
%

For comparison purposes, we also consider three variants of the above formulation based on three alternative concave approximations of $l_{0}$.
In particular, we test the two step function approximations proposed in \cite{rinaldi2010feature}, namely $(u+\varepsilon )^{q}$ where $0 < q < 1$ and $\varepsilon>0$, $- \frac{1}{(u+\varepsilon )^q}$ with $q \geq 1$ and $\varepsilon>0$, and the approximation  $\ln(u+\varepsilon)$ for $u \geq 0$, where $\varepsilon>0$ \cite{weston2003use}. These approximations will be referred to as, respectively, $appr_{1}$, $appr_{2}$ and $log$.

From now on, the formulation proposed in \cite{blanquero2020sparsity} with the objective function \eqref{eq:sparse-orct-blanquero} where $\lambda^G = 0$ and $\lambda^L = 0$ will be referred to as $L_1$ and, respectively, $L_\infty$. While formulation \eqref{eq:l_0-base-formulation} with $\lambda_0^G = 0$ and $\lambda_0^L = 0$ will be referred to as $L_0^{loc}$ and, respectively, $L_0^{glob}$.

\section{Computational results}
\label{sec:comp_results}

In this section we evaluate the testing accuracy and sparsity of the MRCTs 
obtained via concave approximations of the $l_0$  ``norm" 
on 24 datasets from the literature, and compare them with those of the ORCTs found via $l_{1}$ and $l_{\infty}$ regularization as proposed in \cite{blanquero2020sparsity}. 
All the formulations are constructed using Pyomo optimization modeling language in Python 3.6. 
Since we deal with nonlinear nonconvex continuous constrained optimization problems, we adopt the IPOPT 3.11.1 \cite{wachter2006implementation} solver as in \cite{blanquero2020sparsity} and a multistart approach with 10 restarts from different random initial solutions. 
The experiments are carried out on a server with 24 processors Intel(R) Xeon(R) CPU e5645 @2.40GHz 16 GB of RAM.


The section is organized as follows. 
After mentioning dataset information and describing the experimental setup, we report and discuss the results obtained when inducing local and global sparsity separately.
Then 
we summarize the results obtained when both types of sparsity are promoted simultaneously and we conclude with some overall observations.

\begin{table}[h]
\caption{Description of the 24 datasets.}
\label{table:datasets}
\centering
\scalebox{0.5}{%
\begin{tabular}{|lllllc|l|lllllc|}
\cline{1-6} \cline{8-13}
Dataset &
  Abbreviation &
  N &
  p &
  K &
  \multicolumn{1}{l|}{Class distribution} &
   &
  Dataset &
  Abbreviation &
  N &
  p &
  K &
  Class distribution \\ \cline{1-6} \cline{8-13} 
Monks-problems-1 &
  Monks-1 &
  124 &
  11 &
  2 &
  50\%-50\% &
   &
  Iris &
  Iris &
  150 &
  4 &
  3 &
  33.3\%-33.3\%-33.3\% \\
Monks-problems-2 &
  Monks-2 &
  169 &
  11 &
  2 &
  62\%-38\% &
   &
  Hayes-roth &
  Hayes-roth &
  160 &
  15 &
  3 &
  41\%-40\%-19\% \\
Monks-problems-3 &
  Monks-3 &
  122 &
  11 &
  2 &
  51\%-49\% &
   &
  Wine &
  Wine &
  178 &
  13 &
  3 &
  40\%-33\%-27\% \\
\begin{tabular}[c]{@{}l@{}}Connectionist-\\ bench-sonar\end{tabular} &
  Sonar &
  208 &
  60 &
  2 &
  55\%-45\% &
   &
  Seeds &
  Seeds &
  210 &
  7 &
  3 &
  33.3\%-33.3\%-33.3\% \\
Ionosphere &
  Ionosphere &
  351 &
  34 &
  2 &
  64\%-36\% &
   &
  Balance Scale &
  Balance Scale &
  635 &
  16 &
  3 &
  46\%-46\%-8\% \\
\begin{tabular}[c]{@{}l@{}}Breast-cancer-\\ Wisconsin\end{tabular} &
  Wisconsin &
  569 &
  9 &
  2 &
  63\%-37\% &
   &
  \begin{tabular}[c]{@{}l@{}}Contraceptive-\\ method-choice\end{tabular} &
  Contraceptive &
  1473 &
  21 &
  3 &
  42.7\%-34.7\%-22.6\% \\
Credit-approval &
  Creditapproval &
  653 &
  37 &
  2 &
  55\%-45\% &
   &
  \begin{tabular}[c]{@{}l@{}}Thyroid-disease-\\ ann-thyroid\end{tabular} &
  Thyroid &
  3771 &
  21 &
  3 &
  92.5\%-5\%-2.5\% \\
Pima-indians-diabetes &
  Diabetes &
  768 &
  8 &
  2 &
  65\%-35\% &
   &
  Lymphography &
  Lymphography &
  148 &
  50 &
  4 &
  54.7\%-41.2\%-2.8\%-1.3\% \\
\begin{tabular}[c]{@{}l@{}}Statlog-project-\\ German-credit\end{tabular} &
  Germancredit &
  1000 &
  48 &
  2 &
  70\%-30\% &
   &
  Vehicle-silhouettes &
  Vehicle &
  846 &
  18 &
  4 &
  25.7\%-25.6\%-25\%-23.3\% \\
\begin{tabular}[c]{@{}l@{}}Banknote-\\ autothentification\end{tabular} &
  Banknote &
  1372 &
  4 &
  2 &
  56\%-44\% &
   &
  Car-evaluation &
  Car &
  1728 &
  15 &
  4 &
  70\%-22\%-4\%-4\% \\
\begin{tabular}[c]{@{}l@{}}Ozone-level-\\ detection-one\end{tabular} &
  Ozone &
  1848 &
  72 &
  2 &
  97\%-3\% &
   &
  Dermatology &
  Dermatology &
  358 &
  34 &
  6 &
  31\%-19.8\%-16.7\%-13.4\%-13.4\%-5.7\% \\
Spambase &
  Spambase &
  4601 &
  57 &
  2 &
  61\%-39\% &
   &
  Ecoli &
  Ecoli &
  336 &
  7 &
  8 &
  42.5\%-23\%-15.5\%-10.4\%-5.9\%-1.5\%-0.6\%-0.6\% \\ \cline{1-6} \cline{8-13} 
\end{tabular}
}
\end{table}


\subsection{Datasets and experiments} 
\label{Datasets and experiments}

In the experiments, we consider all the datasets from the UCI Machine Learning repository \cite{asuncion2007uci} used in \cite{blanquero2020sparsity} as well as 6
well-known datasets from the KEEL repository \cite{alcala2011keel}. The purpose is to include also datasets with a larger number of features and classes. 
Table \ref{table:datasets} reports the characteristics of the 24 datasets. 
Since the number of classes ranges from 2 to 8, the classification trees are of depth $D=1,2,3$.
Note that for datasets with two classes ($D=1$) the regularizations for local and global sparsity coincide.

The testing accuracy and sparsity of the MRCTs trained with the various formulations is evaluated by means of $k$-fold cross-validation, with $k= 5$. 
Each dataset is randomly divided into $k$  
subsets of samples.  
In turn, every subset of samples is considered as testing set, while the rest is used to train the model. Therefore, all the samples are used both for training and testing.
Then the model accuracy is computed as the average of the accuracies obtained over all $k$ folds. 

For each fold the model is trained from $10$ different random starting solutions. The accuracy of each fold is the average of the accuracies of the $10$ trained models.
As 
in \cite{blanquero2020sparsity}, the following two sparsity indices are considered. The local sparsity, denoted by $\delta^{L}$, is the percentage of predictor variables not used per branch node:
$$\delta ^{L}=\frac{1}{\left | \tau_{B} \right |}\sum_{t\in \tau_{B}}\frac{\left | \left \{ a_{jt}= 0,j=1,\ldots,p \right \} \right |}{p}\times 100.$$
The global sparsity, denoted by $\delta^{G}$, is the percentage of predictor variables not used across the whole tree:
$$\delta ^{G}=\frac{\left | \left \{ a_{j} = 0,j=1,\ldots,p \right \} \right |}{p}\times 100.$$

For the misclassification costs $w_{y_{i} k}$ and the CDF parameter $\gamma$ we took the values as in \cite{blanquero2020sparsity} (if $y_{i}\neq k$ then $w_{y_{i} k}= 0.5$, else $w_{y_{i} k}= 0$, and $\gamma$ = 512), while for the parameters of the concave approximations of the step function we set $q = 1$ and $\varepsilon$ = 10e-6. Moreover, in all our experiments we use the minimum correct classification rate constraint, as defined in \cite{blanquero2018optimal,blanquero2020sparsity} 
with the same parameter settings (for each class a minimum percentage of correctly classified data points equal to $10\%$).


Concerning the equality constraints \eqref{eq:prob-constr}, two implementation strategies are possible: either (i) they are explicitly 
included in the formulation, or (ii) the $P_{it}$ terms in the objective function are replaced with the corresponding right-hand side
of \eqref{eq:prob-constr}.
Preliminary tests showed that, with the adopted optimization solver (the same as in \cite{blanquero2020sparsity}), strategy (i) tends to be substantially more robust with respect to the starting 
solutions, i.e., it sharply reduces 
the frequency with which the solver gets stuck in poor quality solutions nearby the starting ones.
Nonetheless, it affects the computational times: 
with strategy (ii) the training times of all the models are comparable with the ones reported in \cite{blanquero2020sparsity}, 
while strategy (i) 
leads to significantly higher training times. Since our experimental campaign 
focuses on the accuracy and sparsity levels of the trained models, strategy (i) has been chosen. 
Then, for instance,
for the best performing models $L_{1}$ and $L_{0}^{glob}$ and for trees of depth 1 the computational times range from 1.77 ($L_{0}^{glob}$) and 1.63 ($L_{1}$) seconds for Monks-1 to 602.4 ($L_{0}^{glob}$) and 713.3 ($L_{1}$) seconds for Spambase. For trees of depth 2 the computational times range from 13.8 ($L_{0}^{glob}$) and 14.6 ($L_{1}$) seconds for Iris to 1102.5 ($L_{0}^{glob}$) and 1566.3 ($L_{1}$) seconds for Thyroid. For trees of depth 3 the computational times are 3236.9 ($L_{0}^{glob}$) and 3075.1 ($L_{1}$) seconds for Ecoli. 

\subsection{Results for separate local or global sparsity} 
\label{Numerical results on sparsity}

This section is devoted to the comparative results 
for the 24 datasets when local and global sparsity are considered separately.
Tables \ref{table:table2} and \ref{table:table3} report the best results in terms of accuracy and sparsity obtained by each kind of model by varying the regularization parameters ($\lambda^L$ or $\lambda^G$) in the interval $ \left \{ 2^{r}  :\;\;-8\leq r\leq 5,\;\; r\in \mathbb{Z} \right \}.$
As expected, for all the datasets and for both local or global sparsity, when the associated $\lambda$ grows the corresponding $\delta$ index increases.
Moreover, for all the datasets, the sparsest tree (in local or global terms) is obtained when the corresponding $\lambda$ assumes the largest value, although it often has the worst accuracy. 
For all the regularization terms the best 
accuracy levels are often obtained for sparse models where some of the predictor variables are neglected.

Table \ref{table:table2} summarizes the results for the 12 datasets with two classes and for the regularized models $L_{1}$, $L_{\infty}$, $L_{0}^{loc}$, $L_{0}^{glob}$, $appr_{1}$. The results obtained with the $appr_{2}$ and $log$ approximations are not reported because they are outperformed by those obtained with the other $l_0$ "norm" approximations. For these datasets, even if $L_{0}^{loc}$ and $L_{0}^{glob}$ perform almost always better than $appr_{1}$, the latter turns out to be comparable and in few cases also slightly preferable.
Recall that for two-class datasets local and global sparsity coincide ($L_{0}^{loc}$ and $L_{0}^{glob}$ as well as $L_{1}$ and $L_{\infty}$) since $D = 1$ and the trees contain a single branch node. As a consequence, the results are very similar and the slight differences are accounted for by the different initial solutions. 

According to the comparative results we can distinguish three cases. We describe two representative examples for each case. \\
\indent For a first group of datasets, which includes Wisconsin, Monks-2, Monks-3 and Ozone, 
the accuracy obtained via $L_{0}^{loc}$ and $L_{0}^{glob}$ is slightly lower (never by more than 0.6\%) compared to 
the regularizations in \cite{blanquero2020sparsity},
but for all these dataset except Monks-3 great gain in sparsity is achieved, by often removing twice as many predictor variables.  
For Wisconsin the best $l_0$-based model ($L_{0}^{glob}$) reaches an accuracy of 96.1\% by removing 48.89\% of the features; while the best $l_1$-$l_\infty$ model ($L_1$) achieves an accuracy of 96.5\% by removing 24.45\% of the predictor variables. For Monks-3 the best $l_0$-based model ($appr_{1}$) reaches an accuracy of 93.9\% by removing 82.36\% of the features, while 
the best $l_1$-$l_\infty$ model (both $L_1$ and $L_{\infty}$) achieves an accuracy of 93.5\% by removing 88.24\% of the predictor variables.   
%

For a second group of datasets, including Diabetes, Monks-1, Creditapproval, Germancredit and Spambase, 
the approximate $l_0$ regularizations are, in terms of accuracy, at least as good as the previous ones and a higher gain in sparsity is obtained. 
For Monks-1, the best $l_0$-based model ($L_{0}^{glob}$) reaches an accuracy of 87.8\% by removing 48.83\% of the features; while the  $L_{1}$ yields an accuracy of 87.6\% by removing 17.29\% of the predictor variables and the $L_{\infty}$ reaches an accuracy of 87.4\% by removing 45.53\% of the features. For Spambase the best $l_0$-based model ($L_{0}^{glob}$) and the best $l_1$-$l_\infty$ model  ($L_{\infty}$) reach both an accuracy of 89.9\%, but the former removes 4.22\% of the features, while the latter $1.97\%$. 

For a third group of datasets, consisting of Sonar and Ionosphere, the $l_1$-$l_\infty$ regularizations compare favourably with the approximate $l_0$ regularizations in terms of accuracy and sparsity. 
For Sonar the best $l_0$-based model ($L_{0}^{glob}$) reaches an accuracy of 74.2\% by removing 18.98\% of the features; while the best $l_1$-$l_\infty$ model ($L_1$) achieves an accuracy of 76.2\% by removing 27.14\% of the predictor variables. In the case of Ionosphere the best $l_0$-based model ($L_{0}^{loc}$) reaches an accuracy of 86.51\% when removing 3.36\% of the features; while the best $l_1$-$l_\infty$ model ($L_\infty$)  achieves an accuracy of 88\% by removing 84.94\% of the predictor variables.
Finally, note that for the Banknote datatset all regularizations yield MRCTs with the same accuracy and lead to no sparsification.

The above results for two-class datasets indicate that the $l_0$-based and the the $l_1$-$l_\infty$-based models are comparable, with the $l_0$-based often superior in terms of solution sparsity except for two datasets.

\begin{figure}[h]
\centering
\includegraphics[scale=0.8]{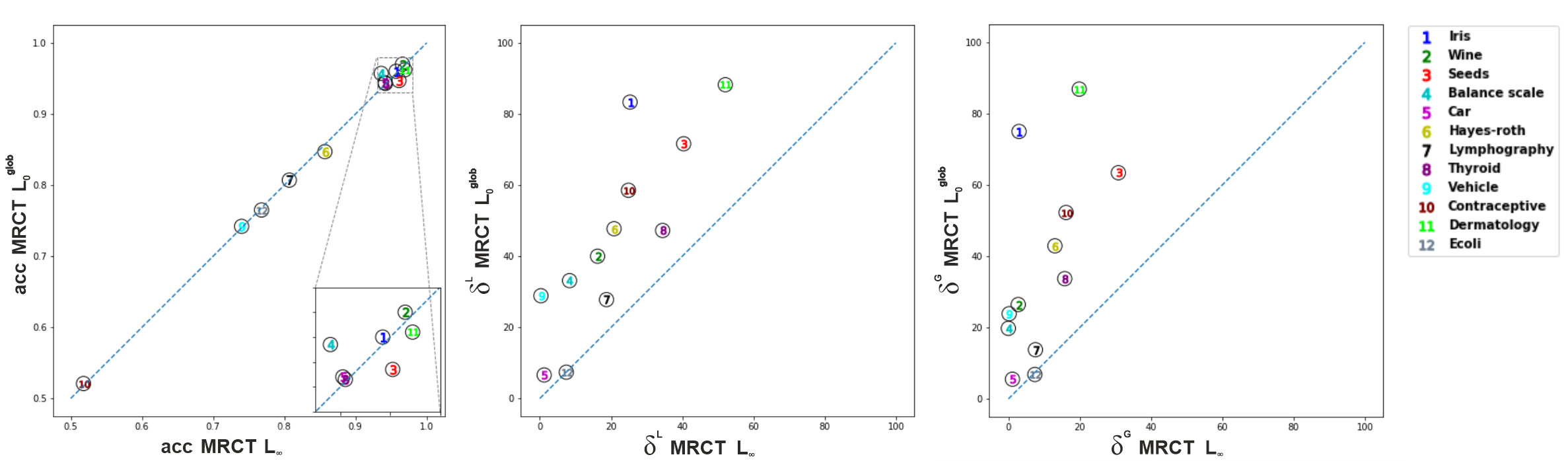}
\caption{\label{fig:cas-sc-template} Comparison of the global regularization proposals for datasets with K $>$ 2 classes in terms of accuracy on the left side, local sparsity in the middle and global sparsity on the right side. On the y-axis the $L_{0}^{glob}$ model and on x-axis the $L_{\infty}$ one. }

\centering
\includegraphics[scale=0.8]{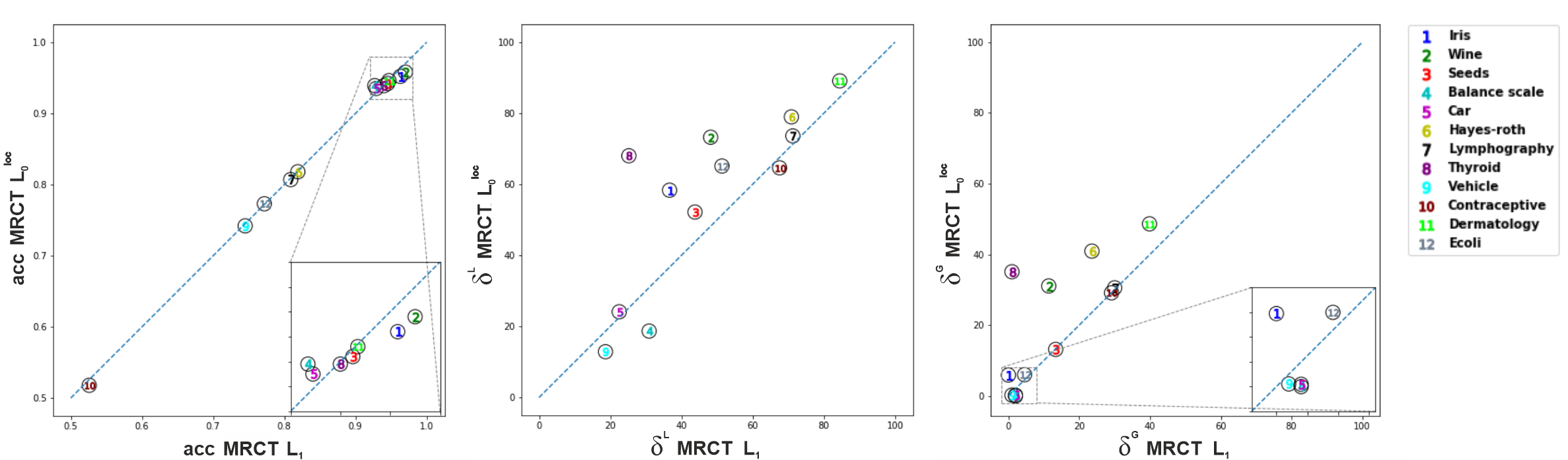}
\caption{\label{fig:cas-sc-template2} Comparison of the local regularization proposals for datasets with K $>$ 2 classes in terms of accuracy on the left side, local sparsity in the middle and global sparsity on the right side. On the y-axis the $L_{0}^{loc}$ model and on x-axis the $L_{1}$ one. }
\end{figure}

Table \ref{table:table3} summarizes the results for 12 multi-class datasets with $K \geq 3$ classes and for the $L_{1}$, $L_{\infty}$, $L_{0}^{loc}$ and $L_{0}^{glob}$ models ($appr_{1}$ is not reported since it is outperformed by the other $l_0$-based alternatives). 
For such datasets, MRCTs with more than one branch node ($D \geq 2$) are required, and global and local sparsity (associated to the measures $\delta^{G}$ and $\delta^{L}$) clearly differ. 
The plots in Figures \ref{fig:cas-sc-template} and \ref{fig:cas-sc-template2} elaborate the results of Table \ref{table:table3} and allow the comparison, in terms of accuracy, $\delta^{G}$ and $\delta^{L}$, of the regularizations for local and global sparsity. 

Figure \ref{fig:cas-sc-template} shows the comparison between the global models $L_{0}^{glob}$ and $L_{\infty}$, while Figure \ref{fig:cas-sc-template2} compares the local models $L_{0}^{loc}$ and $L_{1}$. Both figures report, from left to right, the accuracy, the local sparsity level $\delta^{L}$ and the global sparsity $\delta^{L}$. The values for the $l_0$-based models are plotted on the y-axis while those for the $l_1$-$l_\infty$-based models on the x-axis, and each dataset is represented by a circle with an identification number. 
For datasets with $K \geq 3$ classes, both figures show that the $l_0$-based regularizations lead to higher sparsity while guaranteeing comparable accuracy. 
In particular, regarding 
the global sparsity regularizations, the area above the bisectors of the middle and left plots in Figure \ref{fig:cas-sc-template} show that even if the $L_{0}^{glob}$ and $L_{\infty}$ yield comparable accuracy (the circles associated to all datasets lie on around the bisector of the left plot), 
the former regularization is definitely preferable in terms of the sparsity indices $\delta^{L}$ and $\delta^{G}$.
Concerning the local sparsity regularizations, the left plot of Figure \ref{fig:cas-sc-template2} shows that $L_{0}^{loc}$ and $L_{1}$ lead to comparable accuracy 
and the middle and right plots of Figure \ref{fig:cas-sc-template2} show that the $L_{0}^{loc}$ turns out to be often better in terms of $\delta^{L}$, while it is almost always better in terms of $\delta^{G}$. 

\subsection{Results for combined local and global sparsity} 
\label{Results for local and global sparsity}

In this section both  local and global sparsity are simultaneously induced. A grid of $14 \times 14$ pairs of values is considered for the parameters  $(\lambda^{L}, \lambda^{G})$, where
both $\lambda^{G}$ and $\lambda^{L}$ take their values in $ \left \{ 2^{r},\;\;-8\leq r\leq 5\;\;r\in \mathbb{Z} \right \}.$ 
In order to speed up the grid search, for a given pair $(\lambda^{L},\lambda^{G})$, the 10 solutions of the previous pair are adopted as starting solutions. 
The experiments are focused on datasets with $K=3$, $4$, $6$ and $8$, for which sparse MRCTs of depth $D=2$ and $3$ are trained. 
The results 
are presented by means of heatmaps. In Figure \ref{fig:heat} we select two representative examples, all others are shown in the \hyperlink{appendix}{Appendix}.
As in \cite{blanquero2020sparsity}, for each dataset three heatmaps are reported. They respectively represent the average testing accuracy \emph{acc}, the average local sparsity $\delta^{L}$ and the average global sparsity $\delta^{G}$ (over the $10$ runs), as a function of the parameters $\lambda^{G}$ and $\lambda^{L}$. 
The color range of each heatmap goes from dark red to white, where the white corresponds to, respectively, the maximum accuracy and the maximum local or global sparsity achieved. In general, we observe that the best level of accuracy is not always achieved when $(\lambda^{G},\lambda^{L})$ assume the minimum values. 
When comparing the two alternative regularizations, we note that as the values of $\lambda^{G}$ and $\lambda^{L}$ vary the behaviour in terms of accuracy is very similar. 
Comparing the heatmaps we notice that, as the values of the $\lambda^{L}$ and $\lambda^{G}$ regularizations increase, the accuracy decreases faster for the $l_0$-based regularizations than for the $l_{1}$ and $l_{\infty}$. As also shown in \cite{blanquero2020sparsity}, focusing on global sparsity, in general for a fixed $\lambda^{L}$, $\delta^{G}$ has a growing trend and the same behaviour can be observed for $\delta^{L}$ when $\lambda^{G}$ is fixed. As  expected, the gain in $\delta^{L}$ is greater than the gain in $\delta^{G}$ when the $\lambda^{G}$ value changes.

\subsection{Overall observations}
The above experimental results lead to the three following observations. First, for datasets with two classes, the $l_{0}$-based regularization can improve both local and global sparsity, without compromising the classification accuracy. Indeed, most of the times it is comparable to $l_{1}$ regularization in terms of accuracy and often better in terms of sparsity. Second, for datasets with more than two classes, the $l_{0}$-based models for global sparsity, that is $L^{glob}_{0}$, are almost always the best one in terms of both accuracy and sparsity. Third, when both local and global regularization terms are simultaneously considered the $l_{0}$-based ones are comparable with the combined $l_{1}$ and $l_{\infty}$-based ones.

\begin{sidewaystable}[ph!]

\caption{Results of 
the main sparsification 
methods for MRCTs of depth $D = 1$. The best result in terms of out-of-sample accuracy as a function of $\lambda$ and the respective local sparsity $\delta^{L}$ are reported for each method.}
\label{table:table2}
\centering
\resizebox{\columnwidth}{!}{%
\fontsize{9.5pt}{10.25pt}\selectfont
\begin{tabular}{llccclccclccclccclccclccc}
\hline
 &
   &
  \multicolumn{3}{c}{Monks1} &
   &
  \multicolumn{3}{c}{Monks2} &
   &
  \multicolumn{3}{c}{Monks3} &
   &
  \multicolumn{3}{c}{Sonar} &
   &
  \multicolumn{3}{c}{Ionosphere} &
   &
  \multicolumn{3}{c}{Wisconsin} \\ \cline{3-5} \cline{7-9} \cline{11-13} \cline{15-17} \cline{19-21} \cline{23-25} 
 &
   &
  \multicolumn{1}{l}{} &
  \multicolumn{1}{l}{} &
  \multicolumn{1}{l}{} &
   &
  \multicolumn{1}{l}{} &
  \multicolumn{1}{l}{} &
  \multicolumn{1}{l}{} &
   &
  \multicolumn{1}{l}{} &
  \multicolumn{1}{l}{} &
  \multicolumn{1}{l}{} &
   &
  \multicolumn{1}{l}{} &
  \multicolumn{1}{l}{} &
  \multicolumn{1}{l}{} &
   &
  \multicolumn{1}{l}{} &
  \multicolumn{1}{l}{} &
  \multicolumn{1}{l}{} &
   &
  \multicolumn{1}{l}{} &
  \multicolumn{1}{l}{} &
  \multicolumn{1}{l}{} \\
\multicolumn{1}{c}{Regularized model} &
   &
  \multicolumn{1}{l}{Acc.} &
  \multicolumn{1}{l}{$\delta^{L}$=$\delta^{G}$} &
  \multicolumn{1}{l}{$\lambda$} &
   &
  \multicolumn{1}{l}{Acc.} &
  \multicolumn{1}{l}{$\delta^{L}$=$\delta^{G}$} &
  \multicolumn{1}{l}{$\lambda$} &
   &
  \multicolumn{1}{l}{Acc.} &
  \multicolumn{1}{l}{$\delta^{L}$=$\delta^{G}$} &
  \multicolumn{1}{l}{$\lambda$} &
   &
  \multicolumn{1}{l}{Acc.} &
  \multicolumn{1}{l}{$\delta^{L}$=$\delta^{G}$} &
  \multicolumn{1}{l}{$\lambda$} &
   &
  \multicolumn{1}{l}{Acc.} &
  \multicolumn{1}{l}{$\delta^{L}$=$\delta^{G}$} &
  \multicolumn{1}{l}{$\lambda$} &
   &
  \multicolumn{1}{l}{Acc.} &
  \multicolumn{1}{l}{$\delta^{L}$=$\delta^{G}$} &
  \multicolumn{1}{l}{$\lambda$} \\ \hline
$L_{1}$ &
   &
  0.876 &
  17.29 &
  $2^{-8}$ &
   &
  0.728 &
  26.24 &
  $2^{-4}$ &
   &
  0.935 &
  88.24 &
  $2^{0}$ &
   &
  0.762 &
  27.14 &
  $2^{-2}$ &
   &
  0.879 &
  85.06 &
  $2^{3}$ &
   &
  0.965 &
  24.45 &
  $2^{4}$ \\
$L_{\infty}$ &
   &
  0.874 &
  45.53 &
  $2^{-3}$ &
   &
  0.728 &
  26.7 &
  $2^{-3}$ &
   &
  0.935 &
  88.24 &
  $2^{1}$ &
   &
  0.759 &
  26.94 &
  $2^{-2}$ &
   &
  0.88 &
  84.94 &
  $2^{3}$ &
   &
  0.964 &
  31.1 &
  $2^{5}$ \\
$L_{0}^{loc}$ &
   &
  0.874 &
  46.24 &
  $2^{-8}$ &
   &
  0.717 &
  55.16 &
  $2^{-6}$ &
   &
  0.933 &
  85.53 &
  $2^{-5}$ &
   &
  0.736 &
  28.14 &
  $2^{-4}$ &
   &
  0.865 &
  3.36 &
  $2^{-8}$ &
   &
  0.961 &
  48.89 &
  $2^{2}$ \\
$L_{0}^{glob}$ &
   &
  0.878 &
  48.83 &
  $2^{-7}$ &
   &
  0.722 &
  47.18 &
  $2^{-8}$ &
   &
  0.934 &
  82.7 &
  $2^{-5}$ &
   &
  0.742 &
  18.98 &
  $2^{-5}$ &
   &
  0.863 &
  3.94 &
  $2^{-7}$ &
   &
  0.961 &
  48.89 &
  $2^{2}$ \\
  $appr_1$ &
   &
  0.853 &
  55.09 &
  $2^{-8}$ &
   &
  0.718 &
  53.77 &
  $2^{-8}$ &
   &
  0.939 &
  82.36 &
  $2^{-5}$ &
   &
  0.746 &
  20 &
  $2^{-5}$ &
   &
  0.861 &
  60.71 &
  $2^{-2}$ &
   &
  0.961 &
  45.1 &
  $2^{0}$ \\ \hline
 &
   &
  \multicolumn{1}{l}{} &
  \multicolumn{1}{l}{} &
  \multicolumn{1}{l}{} &
   &
  \multicolumn{1}{l}{} &
  \multicolumn{1}{l}{} &
  \multicolumn{1}{l}{} &
   &
  \multicolumn{1}{l}{} &
  \multicolumn{1}{l}{} &
  \multicolumn{1}{l}{} &
   &
  \multicolumn{1}{l}{} &
  \multicolumn{1}{l}{} &
  \multicolumn{1}{l}{} &
   &
  \multicolumn{1}{l}{} &
  \multicolumn{1}{l}{} &
  \multicolumn{1}{l}{} &
   &
  \multicolumn{1}{l}{} &
  \multicolumn{1}{l}{} &
  \multicolumn{1}{l}{} \\
 &
   &
  \multicolumn{1}{l}{} &
  \multicolumn{1}{l}{} &
  \multicolumn{1}{l}{} &
   &
  \multicolumn{1}{l}{} &
  \multicolumn{1}{l}{} &
  \multicolumn{1}{l}{} &
   &
  \multicolumn{1}{l}{} &
  \multicolumn{1}{l}{} &
  \multicolumn{1}{l}{} &
   &
  \multicolumn{1}{l}{} &
  \multicolumn{1}{l}{} &
  \multicolumn{1}{l}{} &
   &
  \multicolumn{1}{l}{} &
  \multicolumn{1}{l}{} &
  \multicolumn{1}{l}{} &
   &
  \multicolumn{1}{l}{} &
  \multicolumn{1}{l}{} &
  \multicolumn{1}{l}{} \\ \hline
 &
   &
  \multicolumn{3}{c}{Creditapproval} &
   &
  \multicolumn{3}{c}{Diabetes} &
   &
  \multicolumn{3}{c}{Germancredit} &
   &
  \multicolumn{3}{c}{Banknote} &
   &
  \multicolumn{3}{c}{Ozone} &
   &
  \multicolumn{3}{c}{Spambase} \\ \cline{3-5} \cline{7-9} \cline{11-13} \cline{15-17} \cline{19-21} \cline{23-25} 
 &
   &
  \multicolumn{1}{l}{} &
  \multicolumn{1}{l}{} &
  \multicolumn{1}{l}{} &
   &
  \multicolumn{1}{l}{} &
  \multicolumn{1}{l}{} &
  \multicolumn{1}{l}{} &
   &
  \multicolumn{1}{l}{} &
  \multicolumn{1}{l}{} &
  \multicolumn{1}{l}{} &
   &
  \multicolumn{1}{l}{} &
  \multicolumn{1}{l}{} &
  \multicolumn{1}{l}{} &
   &
  \multicolumn{1}{l}{} &
  \multicolumn{1}{l}{} &
  \multicolumn{1}{l}{} &
   &
  \multicolumn{1}{l}{} &
  \multicolumn{1}{l}{} &
  \multicolumn{1}{l}{} \\
\multicolumn{1}{c}{Regularized model} & 
   &
  \multicolumn{1}{l}{Acc.} &
  \multicolumn{1}{l}{$\delta^{L}$=$\delta^{G}$} &
  \multicolumn{1}{l}{$\lambda$} &
   &
  \multicolumn{1}{l}{Acc.} &
  \multicolumn{1}{l}{$\delta^{L}$=$\delta^{G}$} &
  \multicolumn{1}{l}{$\lambda$} &
   &
  \multicolumn{1}{l}{Acc.} &
  \multicolumn{1}{l}{$\delta^{L}$=$\delta^{G}$} &
  \multicolumn{1}{l}{$\lambda$} &
   &
  \multicolumn{1}{l}{Acc.} &
  \multicolumn{1}{l}{$\delta^{L}$=$\delta^{G}$} &
  \multicolumn{1}{l}{$\lambda$} &
   &
  \multicolumn{1}{l}{Acc.} &
  \multicolumn{1}{l}{$\delta^{L}$=$\delta^{G}$} &
  \multicolumn{1}{l}{$\lambda$} &
   &
  \multicolumn{1}{l}{Acc.} &
  \multicolumn{1}{l}{$\delta^{L}$=$\delta^{G}$} &
  \multicolumn{1}{l}{$\lambda$} \\ \cline{1-5} \cline{7-9} \cline{11-13} \cline{15-17} \cline{19-21} \cline{23-25} 
$L_{1}$ &
   &
  0.862 &
  92.05 &
  $2^{1}$ &
   &
  0.779 &
  0 &
  $2^{-3}$ &
   &
  0.745 &
  5.6 &
  $2^{-7}$ &
   &
  0.993 &
  0 &
  $2^{-8}$ &
   &
  0.969 &
  96.39 &
  $2^{5}$ &
   &
  0.899 &
  0 &
  $2^{-4}$ \\
$L_{\infty}$ &
   &
  0.864 &
  95.66 &
  $2^{3}$ &
   &
  0.779 &
  0 &
  $2^{-4}$ &
   &
  0.745 &
  6.89 &
  $2^{-4}$ &
   &
  0.993 &
  0 &
  $2^{-8}$ &
   &
  0.969 &
  94.45 &
  $2^{5}$ &
   &
  0.899 &
  1.97 &
  $2^{-3}$ \\
$L_{0}^{loc}$ &
   &
  0.864 &
  97.83 &
  $2^{3}$ &
   &
  0.779 &
  5 &
  $2^{-2}$ &
   &
  0.747 &
  9.58 &
  $2^{-7}$ &
   &
  0.993 &
  0 &
  $2^{-8}$ &
   &
  0.969 &
  96 &
  $2^{5}$ &
   &
  0.899 &
  3.37 &
  $2^{-4}$ \\
$L_{0}^{glob}$ &
   &
  0.863 &
  97.74 &
  $2^{1}$ &
   &
  0.779 &
  5 &
  $2^{-2}$ &
   &
  0.745 &
  8.23 &
  $2^{-7}$ &
   &
  0.993 &
  0 &
  $2^{-8}$ &
   &
  0.969 &
  100 &
  $2^{5}$ &
   &
  0.899 &
  4.22 &
  $2^{-3}$ \\
$appr_1$ &
   &
  0.855 &
  41.96 &
  $2^{-5}$ &
   &
  0.777 &
  2.75 &
  $2^{-5}$ &
   &
  0.746 &
  21.58 &
  $2^{-5}$ &
   &
  0.993 &
  0 &
  $2^{-7}$ &
   &
  0.969 &
  100 &
  $2^{5}$ &
   &
  0.899 &
  5.97 &
  $2^{-4}$ \\ \hline
\end{tabular}


}

\caption{Results of 
the sparsification methods 
for MRCTs of depth $D > 1$. The best result in terms of out-of-sample accuracy as a function of $\lambda$ and the respective local $\delta^{L}$ and global $\delta^{G}$ sparsity are reported for each method.}
\label{table:table3}
\centering
\resizebox{\columnwidth}{!}{%
\begin{tabular}{llccclcccclcccclcccclccccccccc}
\hline
  &
  \multicolumn{4}{c}{Iris} &
   &
  \multicolumn{4}{c}{Wine} &
   &
  \multicolumn{4}{c}{Seeds} &
   &
  \multicolumn{4}{c}{Balance scale} &
   &
  \multicolumn{4}{c}{Dermatology} &
   &
  \multicolumn{4}{c}{Ecoli} \\ \cline{2-5} \cline{7-10} \cline{12-15} \cline{17-20} \cline{22-25} \cline{27-30} 
 &
   &
  \multicolumn{1}{l}{} &
  \multicolumn{1}{l}{} &
  \multicolumn{1}{l}{} &
   &
  \multicolumn{1}{l}{} &
  \multicolumn{1}{l}{} &
  \multicolumn{1}{l}{} &
  \multicolumn{1}{l}{} &
   &
  \multicolumn{1}{l}{} &
  \multicolumn{1}{l}{} &
  \multicolumn{1}{l}{} &
  \multicolumn{1}{l}{} &
   &
  \multicolumn{1}{l}{} &
  \multicolumn{1}{l}{} &
  \multicolumn{1}{l}{} &
  \multicolumn{1}{l}{} &
   &
  \multicolumn{1}{l}{} &
  \multicolumn{1}{l}{} &
  \multicolumn{1}{l}{} &
  \multicolumn{1}{l}{} &
   &
  \multicolumn{1}{l}{} &
  \multicolumn{1}{l}{} &
  \multicolumn{1}{l}{} &
  \multicolumn{1}{l}{} \\
Regularized model & 
  Acc. &
  \multicolumn{1}{l}{$\delta^{L}$} &
  \multicolumn{1}{l}{$\delta^{G}$} &
  \multicolumn{1}{l}{$\lambda$} &
   &
  \multicolumn{1}{l}{Acc.} &
  \multicolumn{1}{l}{$\delta^{L}$} &
  \multicolumn{1}{l}{$\delta^{G}$} &
  \multicolumn{1}{l}{$\lambda$} &
   &
  \multicolumn{1}{l}{Acc.} &
  \multicolumn{1}{l}{$\delta^{L}$} &
  \multicolumn{1}{l}{$\delta^{G}$} &
  \multicolumn{1}{l}{$\lambda$} &
   &
  \multicolumn{1}{l}{Acc.} &
  \multicolumn{1}{l}{$\delta^{L}$} &
  \multicolumn{1}{l}{$\delta^{G}$} &
  \multicolumn{1}{l}{$\lambda$} &
   &
  \multicolumn{1}{l}{Acc.} &
  \multicolumn{1}{l}{$\delta^{L}$} &
  \multicolumn{1}{l}{$\delta^{G}$} &
  \multicolumn{1}{l}{$\lambda$} &
   &
  \multicolumn{1}{l}{Acc.} &
  \multicolumn{1}{l}{$\delta^{L}$} &
  \multicolumn{1}{l}{$\delta^{G}$} &
  \multicolumn{1}{l}{$\lambda$} \\ \hline
$L_{1}$ &
  \multicolumn{1}{c}{0.963} &
  36.67 &
  0 &
  $2^{-5}$ &
  \multicolumn{1}{c}{} &
  0.97 &
  48.21 &
  11.39 &
  $2^{-2}$ &
  \multicolumn{1}{c}{} &
  0.945 &
  43.81 &
  13.35 &
  $2^{-1}$ &
  \multicolumn{1}{c}{} &
  0.927 &
  30.94 &
  2 &
  $2^{1}$ &
   &
  0.947 &
  15.53 &
  60.12 &
  $2^{-4}$ &
  \multicolumn{1}{c}{} &
  0.772 &
  51.39 &
  4.57 &
  $2^{0}$ \\
$L_{\infty}$ &
  \multicolumn{1}{c}{0.957} &
  25.33 &
  3 &
  $2^{-3}$ &
  \multicolumn{1}{c}{} &
  0.966 &
  16.21 &
  2.77 &
  $2^{-3}$ &
  \multicolumn{1}{c}{} &
  0.961 &
  40.39 &
  30.86 &
  $2^{3}$ &
  \multicolumn{1}{c}{} &
  0.936 &
  8.34 &
  0 &
  $2^{3}$ &
   &
  0.969 &
  47.9 &
  80.12 &
  $2^{-2}$ &
  \multicolumn{1}{c}{} &
  0.768 &
  7.43 &
  7.43 &
  $2^{2}$ \\
$L_{0}^{loc}$ &
  \multicolumn{1}{c}{0.952} &
  58.34 &
  5.9 &
  $2^{-4}$ &
  \multicolumn{1}{c}{} &
  0.958 &
  73.24 &
  31.08 &
  $2^{-3}$ &
  \multicolumn{1}{c}{} &
  0.942 &
  52.19 &
  13.15 &
  $2^{-3}$ &
  \multicolumn{1}{c}{} &
  0.939 &
  18.7 &
  0 &
  $2^{-3}$ &
   &
  0.946 &
  10.97 &
  51.36 &
  $2^{-7}$ &
  \multicolumn{1}{c}{} &
  0.773 &
  65.14 &
  6 &
  $2^{-1}$ \\
$L_{0}^{glob}$ &
  \multicolumn{1}{c}{0.960} &
  83.33 &
  75 &
  $2^{5}$ &
  \multicolumn{1}{c}{} &
  0.97 &
  40 &
  26.47 &
  $2^{-5}$ &
  \multicolumn{1}{c}{} &
  0.947 &
  71.62 &
  63.43 &
  $2^{2}$ &
  \multicolumn{1}{c}{} &
  0.957 &
  33.14 &
  19.8 &
  $2^{2}$ &
   &
  0.962 &
  11.78 &
  13.13 &
  $2^{-1}$ &
  \multicolumn{1}{c}{} &
  0.765 &
  7.39 &
  6.86 &
  $2^{-1}$ \\ \hline
 &
   &
  \multicolumn{1}{l}{} &
  \multicolumn{1}{l}{} &
  \multicolumn{1}{l}{} &
   &
  \multicolumn{1}{l}{} &
  \multicolumn{1}{l}{} &
  \multicolumn{1}{l}{} &
  \multicolumn{1}{l}{} &
   &
  \multicolumn{1}{l}{} &
  \multicolumn{1}{l}{} &
  \multicolumn{1}{l}{} &
  \multicolumn{1}{l}{} &
   &
  \multicolumn{1}{l}{} &
  \multicolumn{1}{l}{} &
  \multicolumn{1}{l}{} &
  \multicolumn{1}{l}{} &
   &
  \multicolumn{1}{l}{} &
  \multicolumn{1}{l}{} &
  \multicolumn{1}{l}{} &
  \multicolumn{1}{l}{} &
   &
  \multicolumn{1}{l}{} &
  \multicolumn{1}{l}{} &
  \multicolumn{1}{l}{} &
  \multicolumn{1}{l}{} \\
 &
   &
  \multicolumn{1}{l}{} &
  \multicolumn{1}{l}{} &
  \multicolumn{1}{l}{} &
   &
  \multicolumn{1}{l}{} &
  \multicolumn{1}{l}{} &
  \multicolumn{1}{l}{} &
  \multicolumn{1}{l}{} &
   &
  \multicolumn{1}{l}{} &
  \multicolumn{1}{l}{} &
  \multicolumn{1}{l}{} &
  \multicolumn{1}{l}{} &
   &
  \multicolumn{1}{l}{} &
  \multicolumn{1}{l}{} &
  \multicolumn{1}{l}{} &
  \multicolumn{1}{l}{} &
   &
  \multicolumn{1}{l}{} &
  \multicolumn{1}{l}{} &
  \multicolumn{1}{l}{} &
  \multicolumn{1}{l}{} &
   &
  \multicolumn{1}{l}{} &
  \multicolumn{1}{l}{} &
  \multicolumn{1}{l}{} &
  \multicolumn{1}{l}{} \\ \hline
 &
  \multicolumn{4}{c}{Car} &
   &
  \multicolumn{4}{c}{Hayes-Roth} &
   &
  \multicolumn{4}{c}{Lymphography} &
   &
  \multicolumn{4}{c}{Contraceptive} &
   &
  \multicolumn{4}{c}{Vehicle} &
   &
  \multicolumn{4}{c}{Thyroid} \\ \cline{2-5} \cline{7-10} \cline{12-15} \cline{17-20} \cline{22-25} \cline{27-30} 
 &
   &
  \multicolumn{1}{l}{} &
  \multicolumn{1}{l}{} &
  \multicolumn{1}{l}{} &
   &
  \multicolumn{1}{l}{} &
  \multicolumn{1}{l}{} &
  \multicolumn{1}{l}{} &
  \multicolumn{1}{l}{} &
   &
  \multicolumn{1}{l}{} &
  \multicolumn{1}{l}{} &
  \multicolumn{1}{l}{} &
  \multicolumn{1}{l}{} &
   &
  \multicolumn{1}{l}{} &
  \multicolumn{1}{l}{} &
  \multicolumn{1}{l}{} &
  \multicolumn{1}{l}{} &
   &
  \multicolumn{1}{l}{} &
  \multicolumn{1}{l}{} &
  \multicolumn{1}{l}{} &
  \multicolumn{1}{l}{} &
   &
  \multicolumn{1}{l}{} &
  \multicolumn{1}{l}{} &
  \multicolumn{1}{l}{} &
  \multicolumn{1}{l}{} \\
 Regularized model & 
  Acc. &
  \multicolumn{1}{l}{$\delta^{L}$} &
  \multicolumn{1}{l}{$\delta^{G}$} &
  \multicolumn{1}{l}{$\lambda$} &
   &
  \multicolumn{1}{l}{Acc.} &
  \multicolumn{1}{l}{$\delta^{L}$} &
  \multicolumn{1}{l}{$\delta^{G}$} &
  \multicolumn{1}{l}{$\lambda$} &
   &
  \multicolumn{1}{l}{Acc.} &
  \multicolumn{1}{l}{$\delta^{L}$} &
  \multicolumn{1}{l}{$\delta^{G}$} &
  \multicolumn{1}{l}{$\lambda$} &
   &
  \multicolumn{1}{l}{Acc.} &
  \multicolumn{1}{l}{$\delta^{L}$} &
  \multicolumn{1}{l}{$\delta^{G}$} &
  \multicolumn{1}{l}{$\lambda$} &
   &
  \multicolumn{1}{l}{Acc.} &
  \multicolumn{1}{l}{$\delta^{L}$} &
  \multicolumn{1}{l}{$\delta^{G}$} &
  \multicolumn{1}{l}{$\lambda$} &
   &
  \multicolumn{1}{l}{Acc.} &
  \multicolumn{1}{l}{$\delta^{L}$} &
  \multicolumn{1}{l}{$\delta^{G}$} &
  \multicolumn{1}{l}{$\lambda$} \\ \hline
$L_{1}$ &
  0.929 &
  22.54 &
  2 &
  $2^{-7}$ &
   &
  0.819 &
  70.89 &
  23.6 &
  $2^{-3}$ &
   &
  0.809 &
  71.3 &
  30.04 &
  $2^{-2}$ &
   &
  0.526 &
  67.58 &
  29.14 &
  $2^{3}$ &
   &
  0.745 &
  18.67 &
  1 &
  $2^{-1}$ &
  \multicolumn{1}{c}{} &
  0.94 &
  25.24 &
  1 &
  $2^{-7}$ \\
$L_{\infty}$ &
  0.941 &
  1.21 &
  1.15 &
  $2^{0}$ &
   &
  0.857 &
  20.84 &
  13.06 &
  $2^{2}$ &
   &
  0.807 &
  18.72 &
  7.6 &
  $2^{-7}$ &
   &
  0.518 &
  24.89 &
  16.19 &
  $2^{3}$ &
   &
  0.74 &
  0.3 &
  0.22 &
  $2^{-1}$ &
  \multicolumn{1}{c}{} &
  0.942 &
  34.48 &
  15.81 &
  $2^{1}$ \\
$L_{0}^{loc}$ &
  0.935 &
  24.13 &
  0.19 &
  $2^{-8}$ &
   &
  0.818 &
  78.97 &
  40.93 &
  $2^{-2}$ &
   &
  0.807 &
  73.61 &
  30.56 &
  $2^{-6}$ &
   &
  0.518 &
  64.63 &
  29.14 &
  $2^{0}$ &
   &
  0.742 &
  12.89 &
  0.22 &
  $2^{-4}$ &
  \multicolumn{1}{c}{} &
  0.939 &
  68.03 &
  35.09 &
  $2^{-1}$ \\
$L_{0}^{glob}$ &
  0.944 &
  6.61 &
  5.53 &
  $2^{-2}$ &
   &
  0.847 &
  47.73 &
  42.93 &
  $2^{-1}$ &
   &
  0.807 &
  27.8 &
  13.8 &
  $2^{-8}$ &
   &
  0.521 &
  58.54 &
  52.28 &
  $2^{2}$ &
   &
  0.742 &
  28.89 &
  23.89 &
  $2^{0}$ &
  \multicolumn{1}{c}{} &
  0.943 &
  47.24 &
  33.727 &
  $2^{0}$ \\ \hline
\end{tabular}
}
\end{sidewaystable}

\begin{figure}[ph!]
\centering
\includegraphics[width=\textwidth]{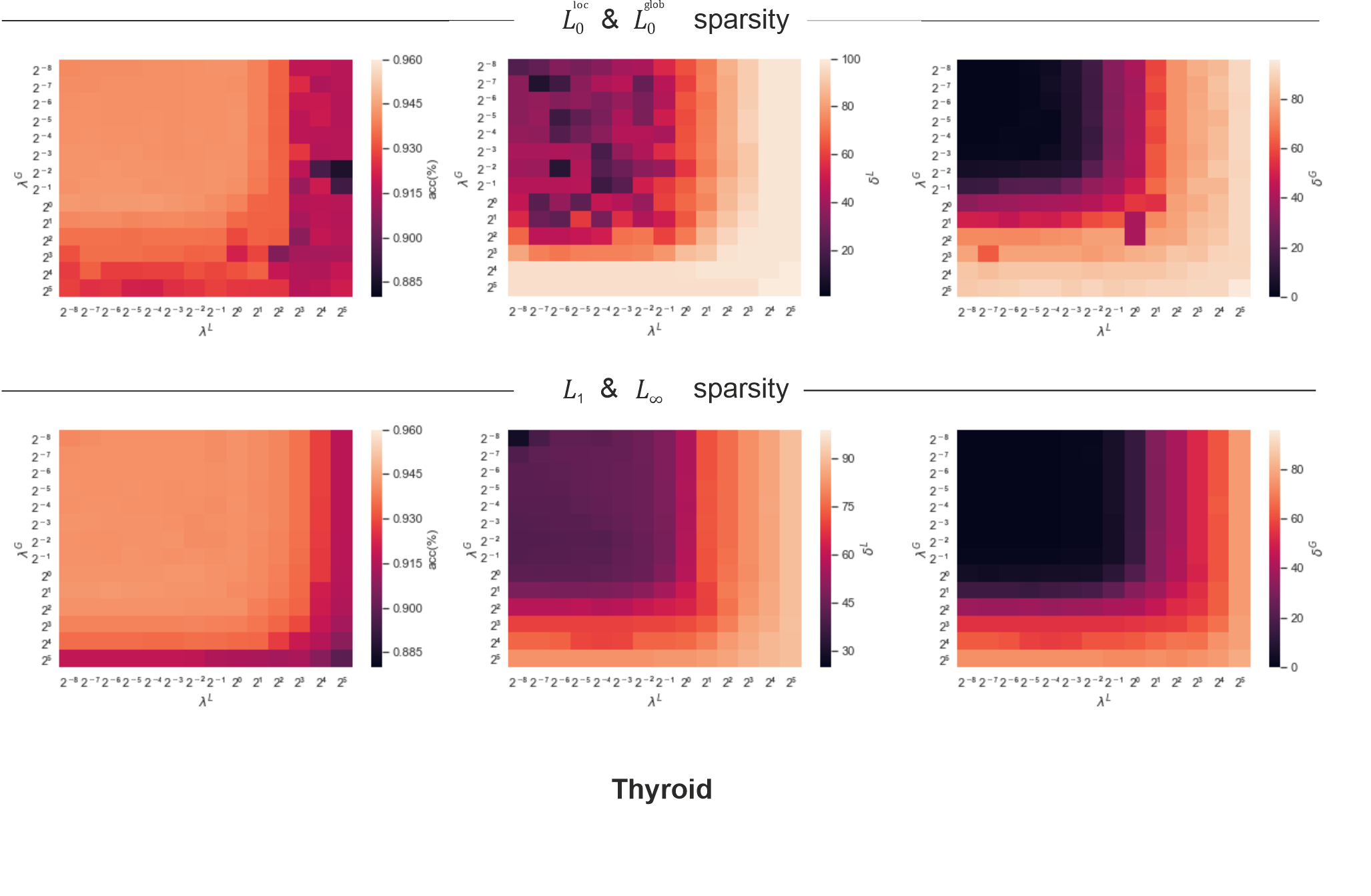}


\centering

\includegraphics[width=\textwidth]{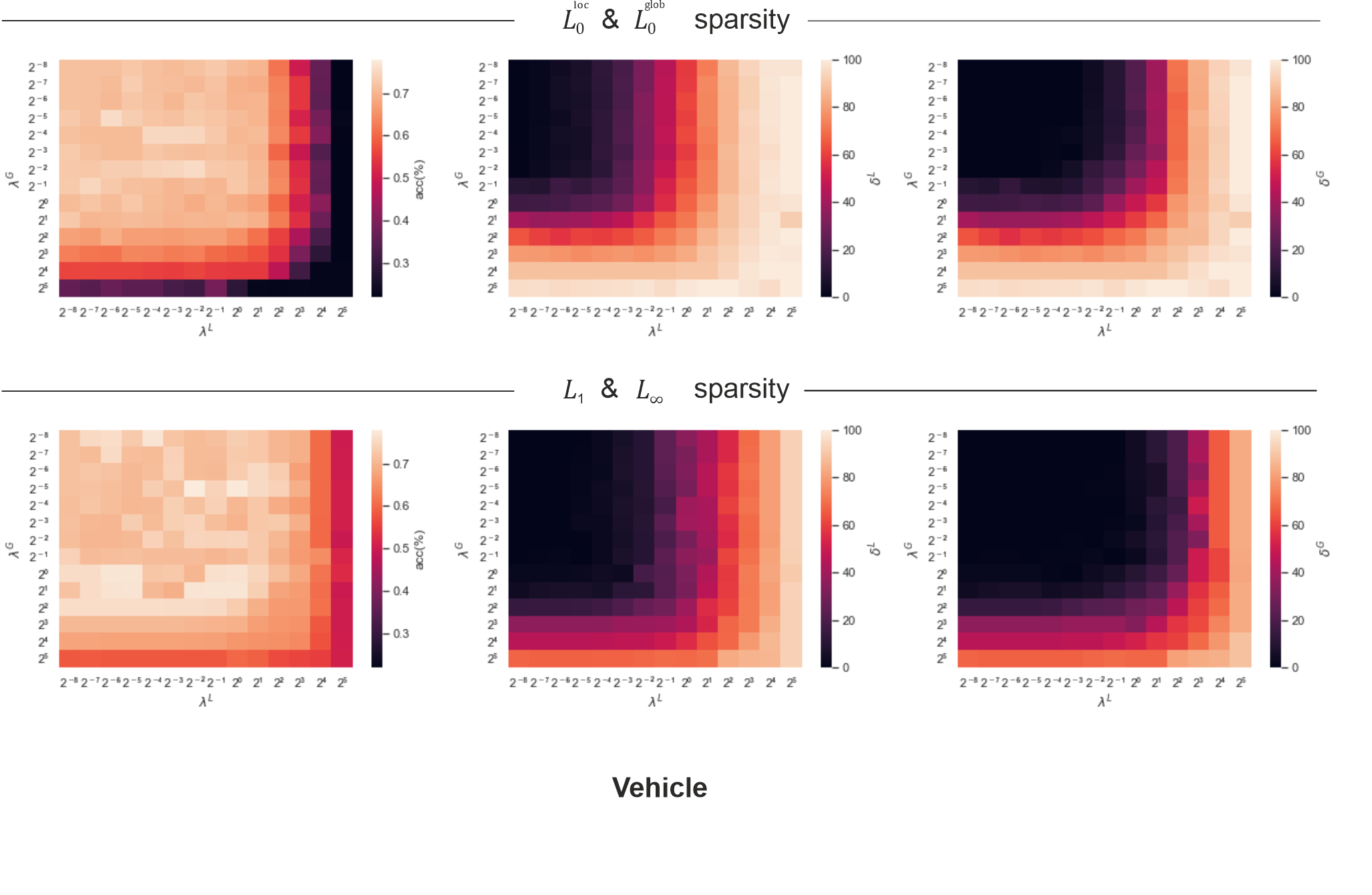}
\caption{\label{fig:heat}Comparison of heatmaps between $l_{0}$-based and $l_{1}-l_{\infty}$ regularizations for Thyroid and Vehicle datasets.}

\end{figure}



	\section{VC dimension of multivariate randomized classification trees}
	\label{sec:VC-dimension}
	
	We consider binary classification tasks with a generic input vector $\mathbf{x} \in {\mathbb R}^p$ and class label $y \in \{0,1\}$. 
	In statistical learning theory, the Vapnik–Chervonenkis (VC) dimension of a binary classification model is a measure that plays a key role in the generalization bounds, that is, in the upper bounds on the test error (see e.g. \cite{anthony1997computational}).
	The VC dimension measures the expressive power and the complexity of  
	the set $H$ of all the functions that can be implemented by the considered classification model. 
	Given a set of hypotheses $H$, 
	the VC dimension of $H$ is defined as the cardinality of the largest number of data points that can be shattered by $H$.
	A set of $l$ data points $\{\mathbf{x}_{1},\ldots,\mathbf{x}_{l} \} \in {\mathbb R}^p$ is said to be \emph{shattered} by $H=\{ h(\mathbf{x},\bm{\alpha})\}$ indexed by a parameter vector $\bm{\alpha}$ if and only if, for every possible assignment of class labels to those $l$ points (every possible dichotomy) there exists at least one function in $H$ that correctly classifies all the data points (is consistent with the dichotomy). In our case, $H$ is the set of all the functions that can be implemented by a given maximal binary and multivariate randomized classification tree of depth $D$ with $p$ inputs, that is, where every branch node at depth smaller or equal to $D-1$ has exactly two child nodes. 

	It is worth recalling that most decision trees in the literature are deterministic.
	To the best of our knowledge, no explicit formula is known for the VC dimension of multivariate or randomized decision trees. However, some bounds and a few exact results are available for special cases of deterministic trees and for a related randomized ML model. 
	In \cite{mansour1997pessimistic} the VC dimension of 
	univariate deterministic decision trees with $\nu$ nodes and $p$ inputs is proved to be between $\Omega$($\nu$) and $O$($\nu \log p$). In \cite{simon1991vapnik} it is shown that the VC dimension of the set of all the Boolean functions on $p$ variables defined by decision trees of rank at most $r$ is $\sum_{k=1}^r {p \choose k}$. 
	In \cite{yildiz2015vc} the author first shows structure-dependent lower bounds for the VC dimension of univariate deterministic decision trees with binary inputs and then extends them to decision trees with $L$ children per node. In \cite{jiang2000vc} the VC dimension of mixture-of-experts architectures with $p$ inputs and $m$ Bernoulli or logistic regression experts is proved to be bounded below by $m$ and above by $O(m^{4}p^{2})$. 
	
	In the remainder of this section, we determine lower and upper bounds on the VC dimension of maximal 
    MRCTs of depth $D$ with $D \geq 1$ and two classes.

	\subsection{Lower bounds} 
	

	We start with a simple observation concerning  
	MRCTs of depth $D=1$, that is, with a single branch node. 	
	Let us recall that the well-known perceptron model (see e.g. \cite{anthony1997computational}) maps a generic $p$-dimensional real input vector $\mathbf{x}$ to the binary output $y(\mathbf{x})=\mathbb{1}_{\mathbb{R}^{+}}(w^{T}\mathbf{x}+b)$, where the parameters $w_j$, with $1 \leq j \leq p$, and $b$ take real values.
	\medskip
	
\noindent	
	\hypertarget{obs1}{}\textbf{Observation 1.} \textit{A MRCT of depth $D=1$ with $p$ real (binary) inputs is as powerful as a perceptron with $p$ real (binary) inputs and hence its VC dimension is equal to $p+1$.}
	\medskip

	Note that a MRCT with $p$ real (binary) inputs and a single branch node coincides with a binary logistic regression model whose response probability conditioned on the input variables is:
	$$\mathbb{P}(y=1 \mid \mathbf{x})= \left \{1+e^{( - \gamma (\bm{\beta}^{T}\mathbf{x}+\beta_0 ) ) }  \right \}^{-1} $$
	with the $p+1$-dimensional parameter vector $(\bm{\beta}^{T},\beta_0)$.
	Considering an appropriate threshold $\rho$ and defining
	$ y(\mathbf{x}) = \mathbb{1}_{\mathbb{R}^{+}}(\mathbb{P}(y=1 \mid \mathbf{x})-\rho)$,
	there is an obvious equivalence with the perceptron model with $n$ inputs.
    For any fixed value of $\rho$ (e.g. $\rho=0.5$), the equation of the separating hyperplane is
 	$$ \frac{1}{1+e^{(- \gamma (\bm{\beta}^{T}\mathbf{x} + \beta_0 )}}-\rho = 0 $$
	and hence
	$$ \bm{\beta}^{T}\mathbf{x} + \beta_0 + \frac{1}{\gamma} ln(\frac{1-\rho}{\rho})= 0.$$
   
	Since the VC dimension of a perceptron with $p$ real (binary)
	inputs is equal to $p+1$ (see e.g. \cite{anthony1997computational}), a MRCT of depth $D=1$ with $p$ real (binary) inputs has the same VC dimension.

	\medskip
	
	
For MRCTs of depth $D=2$, that is, with three branch nodes, we have:
		
	\hypertarget{prop1}{}
	\begin{proposition}
	\label{prop1-vcdim}
		\hypertarget{prop1-vcdim}{}
		\phantomsection\label{proposition1}
		The VC dimension of a maximal MRCT of depth $D=2$ with $p$ real (binary) inputs is at least $2(p+1)$.
	\end{proposition}
\begin{proof}
	To prove the result we need to exhibit a set of $2(p+1)$ points in ${\mathbb R}^p$ which is shattered by a maximal MRCT of depth $D=2$ with $p$ inputs. 
	Since each branch node at depth $2$ can be viewed as a perceptron with $p$ inputs and its VC dimension is $p+1$ even for binary inputs, we show that there exists a set of $p+1$ vertices of the unit hypercube ${\mathbb B}^p=\{0,1\}^p$, denoted by $V_L$, shattered by the left branch node and a set of $p+1$ vertices of ${\mathbb B}^p$, denoted by $V_R$, shattered by the right branch node such that $V_L \cap V_R=\emptyset$ and their union $V_L \cup V_R$ can be shattered by the overall MRCT of depth $D=2$.
	To do so it suffices to verify that there exist values for the parameters $\textbf{a}$ and $\mu$ of the root node which guarantee the separation of points in $V_L$ from those in $V_R$ with a given probability. Indeed, this implies that the root node can forward all the points in $V_L$ to the left branch node and all those in $V_R$ to the right branch node.
	
	For any given dimension $p \geq 2$, we can consider the subset $V_R \subseteq {\mathbb B}^p$ containing the zero vector and the $p$ vectors $\vec e_i$, with $1 \leq i \leq p$, of the canonical base in dimension $p$, and the subset $V_L$ containing the all-ones vector $\vec 1$ and the $p$ vectors $\vec 1 - \vec e_i$, with $1 \leq i \leq p$. Obviously $V_L$ is the complement of $V_R$ and both sets $V_L$ and $V_R$ are full dimensional.
	
	We exhibit values of the parameters $\textbf{a}$ and $\mu$ associated to the root node and of $\gamma$ in the CDF function such that the two sets $V_L$ and $V_R$ turn out to be separable with a given confidence margin. 
	A possible choice for $\textbf{a}$ and $\mu$ is as follows: 
	$$\textbf{a} = \Big(\frac{p-1}{p},\frac{p-1}{p},...,\frac{p-1}{p}\Big)^T\;\; \mbox{ and } \;\; \mu = \frac{1}{p}.$$
	These values guarantee that, given any $p \geq 2$ and threshold $0<\epsilon<0.5$, the probability for every point in $V_R$ to fall to the left of the root node is
	at most $\epsilon$ and the probability for every point in $V_L$ to fall to the right of the root node is at most $\epsilon$.
	Indeed, for any point in $V_R$, the maximum probability to fall 
	to the left is: 
	$$ \frac{1}{1+e^{-\gamma(\frac{p-1}{p^2}-\frac{1}{p})}} = \frac{1}{1+e^{\gamma(\frac{1}{p^2})}},$$ 
	which is at most $\epsilon$ when
	$$\gamma \geq p^2\,ln(\frac{1-\epsilon}{\epsilon}).$$
	Similarly, for any point in $V_L$, the maximum probability to fall 
	to the right is:
	$$1 - \frac{1}{1+e^{-\gamma(\frac{p^2-3p+1}{p^2})}} = \frac{1}{\frac{1}{e^{-\gamma(\frac{p^2-3p+1}{p^2})}}+1},$$
	which is at most $\epsilon$ when
	$$\gamma \geq \frac{p^2}{p^2-3p+1}\,ln(\frac{1-\epsilon}{\epsilon}).$$

	Therefore the MRCT of depth $D=2$ whose root node has the above parameter values is guaranteed to shatter 
		 the $2(p+1)$ points of the set $V_L \cup V_R$ with a high probability.

	Clearly, since we have exhibited binary points the result is also valid for the special case of maximal MRCTs with binary inputs. 

\end{proof}

	
	For maximal MRCTs of depth $D \geq 3$, we have:
	
	\begin{proposition}
	\label{prop2-vcdim}
		\hypertarget{prop2-vcdim}{}
		The VC dimension of a maximal MRCT of depth $D$ with $p$ real (binary) inputs, where $D \geq 3$ and $D \leq p + 2$, is at least $2^{D-1}(p-D+3)$, assuming that $p-D \leq 2^{p-D+1}-3$.
	\end{proposition}
	
	\begin{proof}
	    We use the lower bound in Proposition \ref{proposition1} for a MRCT of depth $D=2$ as well as the following simple extensions of a result and a recursive procedure for univariate deterministic classification trees with binary inputs described in \cite{yildiz2015vc}.  
	    
	    The extended result states that the VC dimension of a maximal MRCT of depth $D \geq 2$ with $p$ real (binary) inputs is at least the sum of the VC dimensions of its left and right subtrees restricted to $p-1$ 
	    inputs. Indeed, by setting for each data point the additional ($p$-th) variable to $0$ or $1$, we can use this variable at the root node to forward the data points to, respectively, the right subtree or the left subtree.
	    The recursive procedure, denoted as {\sc LB-VC}($T$,$p$), takes as input a MRCT $T$ with $p$ real (binary) inputs and depth $D$ where $D \geq 3$ and returns a lower bound on its VC dimension. Let $T_{L}$ and $T_{R}$ denote the left and, respectively,
        right subtrees of $T$. If $T_{L}$ and $T_{R}$ are maximal MRCTs with $D=2$ the procedure returns $2(p+1)$ else
        it returns {\sc LB-VC}($T_{R}$,$p-1$)+{\sc LB-VC}($T_{L}$,$p-1$).
		
		We apply {\sc LB-VC}($T$,$p$) to the maximal MRCT of depth $D$ with $D \geq 3$. 
		For each branch node of depth $D-1$, consider the subtree containing that branch node and its two children (branch nodes) at depth $D$.
		Clearly, there are $2^{D-2}$ such nodes at depth $D-1$ and, according to Proposition \ref{proposition1}, each corresponding subtree contributes by at least $2(p+1-(D-2))$ to the VC dimension of the overall tree. Thus the VC dimension of a maximal MRCT with $p$ inputs and depth $D \geq 3$ is at least $2^{D-2}(2(p-D+3)))=2^{D-1}(p-D+3)$. Since in the case of binary inputs the unit hypercube $\mathbb{B}^{p}$ has $2^{p}$ distinct vertices, we must obviously have $2^{D-1}(p-D+3) \leq 2^{p}$ and hence $p-D\leq 2^{p-D+1}-3$.
	\end{proof}
	Note that the limitation of the above lower bound lies in the fact that one dimension is lost at each depth exceeding $2$ (for this reason we must have $D \leq p+2$).

The lower bounds on the VC dimension of MRCTs in Propositions \ref{prop1-vcdim} and 
\ref{prop2-vcdim}, which depend on the number of inputs $p$ and depth $D$, may be compared with the VC dimensions of other supervised ML models such as the above-mentioned single linear classifiers (e.g. perceptron or Support Vector Machine) with $p$ inputs ($p+1$) or two-layer Feedforward Neural Networks of linear threshold units with $p$ inputs and $u_h$ units in the hidden layer ($pu_h+1$ if $u_h \leq \frac{2^{p+1}}{p^2+p+2}$, see Theorem 6.2 in \cite{anthony2009neural}).

	\subsection{Upper bound}
	
	MRCTs are parametric supervised learning models with a special graph structure, where each branch node is a probabilistic model with a binary outcome based on the input variables, and the leafs are associated to the class labels.
	
	We start with some considerations concerning the topology and properties of MRCTs. Given a tree of depth $D$, let us distinguish the set $\tau_{B}^{bottom}$ of the branch nodes of the last level $D$ facing the leaf nodes, from the set $\tau_{B} \setminus \tau_{B}^{bottom}$ of those at depth smaller or equal to $D-1$.  
	MRCTs can be viewed as a cascade of Bernoulli random variables. 
	For any input vector $\mathbf{x}$ and branch node $t \in \tau_{B}$, the probability $p_{\mathbf{x}t}$ is determined by a logistic CDF, and the cascade leads to a set of  $m=2^{D-1}$ branch nodes $\tau_{B}^{bottom}$. These branch nodes in $\tau_{B}^{bottom}$ with their logistic models forward the input vector $\mathbf{x}$ to the leaf nodes. Notice that, due to the exponential nature of the logistic CDF, for each bottom level node $t'$ in $\tau_{B}^{bottom}$ the cascade of Bernoulli random variables of the branch nodes in $\tau_{B} \setminus \tau_{B}^{bottom}$ expressed by the product: 
	$$ g_{t'}(\mathbf{x}) = 
	\prod_{t_{l}\in N_{L(t')}}p_{\mathbf{x}t_{l}}\;\prod_{t_{r}\in N_{R(t')}}(1-p_{\mathbf{x}t_{r}}) $$ 
	can be expressed as the classical multinomial logistic regression:
	$$g_{t'}(\mathbf{x}) = \frac{e^{\mathbf{u}_{t'}^{T}\mathbf{x} + v_{t'} } }{\sum_{t \in \tau_{B}^{bottom}} e^{ \mathbf{u}_{t}^{T}\mathbf{x} + v_{t} }}$$

	\noindent parameterized by $(\mathbf{u}_{t}^{T}, v_t) \in {\mathbb R}^{p+1}$ for $t \in \tau_{B}^{bottom}$, where $\mathbf{u}_{t}$ are the coefficient vectors and $v_t$ the intercepts.

	The logistic model associated to each one of the $m$ bottom level nodes $t' \in \tau_{B}^{bottom}$ is as follows: 
	$$\pi_{t'}(\mathbf{x}) = \left\{1+e^{( -\beta_{t'}^{T}\mathbf{x} -\beta_{0 t'} )}\right\}^{-1}, $$  parameterized by $(\bm{\beta}_{t'}^{T},\beta_{0 t'}) \in {\mathbb R}^{p+1}$.

	An example of MRCT of depth $D=3$ is shown in the following figure:\\
	\vspace{-10mm}
	
	\begin{figure}[H]
		\centering
		\includegraphics[scale=0.5]{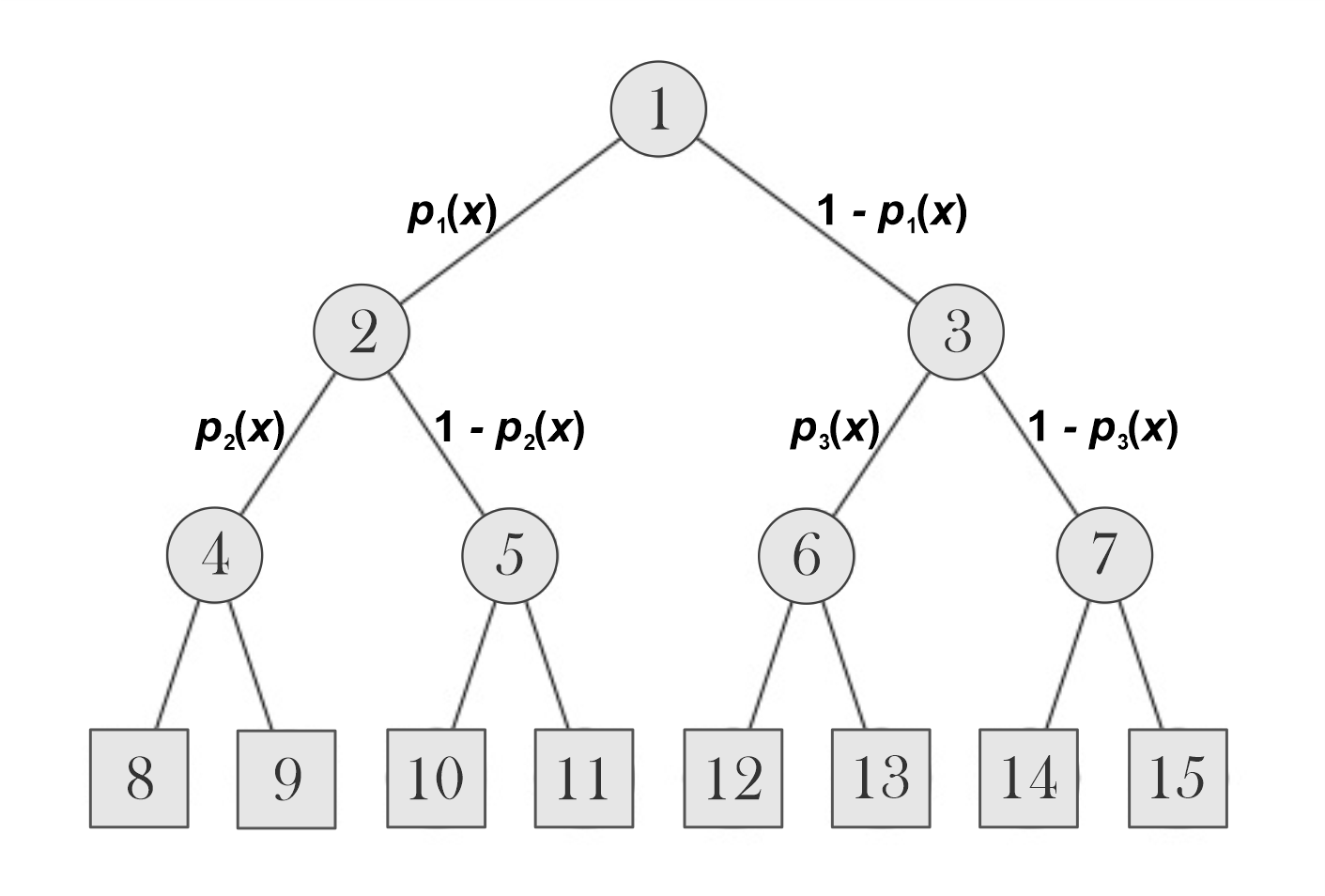}
	\end{figure}
	\noindent where $g_{4}(\mathbf{x}) =p_{1}(\mathbf{x})p_{2}(\mathbf{x}),\, g_{5}(\mathbf{x}) = p_{1}(\mathbf{x})(1 - p_{2}(\mathbf{x}) ),\, g_{6}(\mathbf{x}) =(1 - p_{1}(\mathbf{x})) p_{3}(\mathbf{x}) $ and $ g_{7}(\mathbf{x}) =(1 - p_{1}(\mathbf{x})) (1-p_{3}(\mathbf{x})).$
	\noindent
	For each pair $t \in \tau_L$ and $k \in \{0,1\}$, the variable $c_{kt}$ can be viewed as the probability that the class label $k$ is assigned to leaf node $t$. However, since optimal solutions are integer, we know that a single class label is assigned to each leaf.
	For MRCTs, the probability for a new input vector $\mathbf{x}$ to be assigned 
	to the first class is:
	\hypertarget{eq20}{}
	\begin{equation}
		\mathbb{P}(y=1\mid \mathbf{x}) = \sum_{t \in \tau_{L}}c_{1t}P_{\mathbf{x}t} =\sum_{ t' \in \tau_{B}^{bottom}}\pi_{t'}(\mathbf{x})g_{t'}(\mathbf{x}),
	\end{equation}
	where $P_{\mathbf{x}t}$ is the probability for $\mathbf{x}$ to fall into leaf node $t$.
	Introducing a scalar threshold $\rho$ (e.g $\rho=0.5$), Equation (\hyperlink{eq20}{10}) 
	yields the following discriminant function:
	$$C_{\theta}(\mathbf{x})= \mathbb{1}_{\mathbb{R}^{+}}(\sum_{t \in \tau_{L}}c_{1t}P_{\mathbf{x}t}-\rho),$$ 
	where the parameter vector $\bm{\theta}$ includes all the parameters of the $m$ logistic regressions, at level $D$, $(\bm{\beta}^{T}_{t},\beta_{0 t}), \; t \in \tau_{B}^{bottom}$ and the ones of the $g_{t}(\mathbf{x})$ functions $(\mathbf{u}_{t}^{T},v_{t})$, for $t \in \tau_{B}^{bottom}$, that is $\bm{\theta} \in {\mathbb R}^{2m(p+1)}$.

	Note that Equation (\hyperlink{eq20}{10}) shows the connection between MRCTs and Mixtures of binary Experts (MbEs) \cite{jacobs1991adaptive,jordan1994hierarchical}. MbEs can be seen as a combination of binary experts which implement a probability model of the response conditioned on the input vector. Assuming that we have $m \geq 2$ experts and each one of them has a probability function $\pi_{t'}(\mathbf{x})$, a MbE generates the following conditional probability of belonging to the class label $y = 1$:

	\begin{equation}
		\label{eq:15}
		\qquad \quad p(\mathbf{x}) = \sum_{ t' \in \tau_{B}^{bottom}} \pi_{t'}(\mathbf{x}) g_{t'}(\mathbf{x}) ,
	\end{equation} 
	\noindent
	where $g_{t'}(\mathbf{x})$ are the local weights, called gating functions. 
	
	In \cite{jiang2000vc} the author exploits the result in \cite{karpinski1997polynomial} concerning the VC dimension of neural networks with sigmoidal activation functions to establish an upper bound of $O(m^{4}p^{2})$ on the VC dimension of MbEs with logistic regression models, where $m$ is the number of experts and $p$ the number of inputs. Since in our case $m=2^{D-1}$, we have the following upper bound:
	
	\begin{proposition}
		The VC dimension of a maximal MRCT of depth $D$ with $p$ inputs is at most $O(2^{4(D-1)}p^{2})$.
	\end{proposition}

\section{Decomposition methods for sparse randomized classification trees}
\label{sec:decopmosition}

MRCTs reveal to be promising ML models both in terms of accuracy and of interpretability.
Sparsity enhances interpretability also when the number of features grows. 
However, since the training of sparse MRCTs is formulated as a challenging nonconvex constrained nonlinear optimization problem, the long training times required for larger datasets affect their practical applicability. 
\par
   
In this section, we propose and investigate decomposition methods 
for training good quality MRCTs in significantly shorter computing times. 
First we present a general decomposition scheme, 
then we discuss possible versions of the general algorithm and we propose a specific one whose 
performance is tested on five  
datasets larger than most of those used in Section \ref{Numerical results on sparsity}.

\subsection{A general decomposition scheme}
\label{subsec:general_dec}
    
Decomposition techniques have been extensively considered in the literature for training various learning models such as 
Feedforward Neural Networks 
and Support Vector Machines (e.g., \cite{grippo2015decomposition,huang2006extreme,kingma2014adam,joachims1998making,lucidi2007convergent,chang2011libsvm,manno2018parallel,manno2016convergent}). Indeed, the increasing dimension of the available training sets often 
leads to very challenging large-scale 
optimization problems.\par
Here we devise decomposition methods 
for training sparse MRCTs. As for other ML models, the training problem dimension depends on the size of the training set.
In particular, the number of features determines the number of variables in each branch node, while the number of classes affects 
both the depth of the tree 
and the number of variables in the leaf nodes.\par 
Decomposition methods split the original optimization problem into a sequence of smaller subproblems in which only a subset of variables are optimized at a time, while the remaining ones are kept fixed at their current values.
The set of indices associated to the updated variables is referred to as {\it working set} and it is denoted as 
$W$,  
while its complement is denoted as $\overline W$.\par
The proposed decomposition scheme is designed for the $L_{0}^{glob}$ MRCT model, which turned out to be the most promising one, but everything easily extends to other sparse MRCT models.\par 
Let us rewrite the $L_{0}^{glob}$ formulation with a slightly different notation more suited for a decomposition framework. First, we get rid of the variables $\mu_t$ (the intercept parameters at branch nodes $t \in \tau_{B}$) incorporating them into the variables $a_{jt}$ by simply adding a constant feature with value $-1$ to every 
input vector. 
We denote as $A^+$, $A^-$ the $p \times | \tau_{B} |$ matrices of the auxiliary variables (used to replace the absolute value of the branch nodes variables $a_{jt}$) with elements $a_{jt}^+$ and $a_{jt}^-$ respectively, with $A_{\cdot t}^+=(a_{1t}^+ \ a_{2t}^+ \dots a_{pt}^+)^T$ and $A_{\cdot t}^-=(a_{1t}^- \ a_{2t}^- \dots a_{pt}^-)^T$ as $t$-th columns, and $A_{j \cdot }^+=(a_{j1}^+ \ a_{j2}^+ \dots a_{j\tau_{B}}^+)$ and $A_{j \cdot}^-=(a_{j1}^- \ a_{j2}^- \dots a_{j\tau_{B}}^-)$ as $j$-th rows.
The vector $\bm{\beta}=(\beta_1 \ \beta_2 \dots \beta_p)^T \in {\mathbb R}^p$ includes the upper bounds on the absolute values of the variables $a_{jt}$. Then, we consider the $K \times | \tau_{L} |$ matrix $C$ of leaf node variables with elements $c_{kt}$, with the $t$-th column $\mathbf{C}_{\cdot t}=(c_{1t} \ c_{2t} \dots c_{Kt})^T$ and the $k$-th row $\mathbf{C}_{k \cdot }=(c_{k1} \ c_{k2} \dots c_{k\left\lvert\tau_L\right\rvert})$.\par
         
The objective function of $L_{0}^{glob}$ is the sum of the expected misclassification errors and the sparsity regularization term.
According to the new notation, the error function is written as
         \begin{eqnarray}\label{eq:error}
         \small
             E(A^+,A^-,C)=\sum_{i \in N} \sum_{t \in \tau_{L}} \left[ \prod_{t_l \in N_{L}(t)}F\left(\frac{1}{p} (A_{\cdot t_l}^+-A_{\cdot t_l}^-)^T  \mathbf{x}_i\right) \right. \\
           \left. \prod_{t_r \in N_{R}(t)} 1 -  F\left(
           \frac{1}{p} (A_{\cdot t_r}^+-A_{\cdot t_r}^-)^T \mathbf{x}_i\right)   \sum_{k \in K} w_{y_{i}k} c_{kt}\right], \nonumber
         \end{eqnarray}
while the sparsity regularization term as $S(\bm{\beta})=\sum_{j=1}^{p}(1-e^{-\alpha \beta _{j}}).$ Then the formulation amounts to
\begin{alignat}{2}
     \min_{A^+,A^-,\bm{\beta},C} &  \quad  O(A^+,A^-,\bm{\beta},C)=E(A^+,A^-,C)+\lambda_{0}^G S(\bm{\beta})  \label{pr:formulation} \\
    s.t. \quad & \quad  \sum_{k=1}^{K}c_{kt}=1    \quad t\in \tau_{L}  \nonumber \\
   & \quad \sum_{t \in \tau_{L}}c_{kt}\geq  1  \quad  k=1,\ldots,K \nonumber \\ 
   & \quad  \beta_j \geq a_{jt}^+ + a_{jt}^-  \quad j=1,\dots,p,
    \quad t=1,\dots,\tau_{B}\nonumber \\
    & \quad A^+,A^-\in [0,1]^{p \times | \tau_{B} | }, \;\; \bm{\beta} \in [0,1]^{p}, \;\; C \in [0,1]^{K \times | \tau_{L} | }. & \nonumber
\end{alignat}

Now we are ready to present the proposed decomposition method which is a {\it nodes based} strategy. Indeed, 
at each decomposition step $s$  
a subset of nodes of the tree is selected 
and only the indices of the variables involved in such nodes are inserted in the working set $W^s$.
The latter is composed of 
         \begin{equation}\label{eq:BL}
             \begin{gathered}
                 W_{B}^{s} \subseteq \{1,2,\dots, |\tau_{B} | \} \; \mbox{ and } \;
                 W_{L}^{s} \subseteq \{1,2,\dots, | \tau_{L} | \},
             \end{gathered}
         \end{equation}
i.e., the indices subsets of, respectively, the branch nodes and leaf nodes selected at step $s$,   
with 
         \begin{equation}\label{eq:barBL}
             \begin{gathered}
                 \overline W_{B}^{s} \equiv \{1,2,\dots, | \tau_{B} | \} \setminus W_{B}^{s} \; \mbox{ and } \; 
                 \overline W_{L}^{s} \equiv \{1,2,\dots, | \tau_{L} | \} \setminus W_{L}^{s}
             \end{gathered}
         \end{equation}
as complements. For simplicity, from now on the dependency of the working sets 
on $s$ will be omitted. Let us denote by $A_{W_{B}}^+$ ($A_{W_{B}}^-$) the submatrix of $A^+$ ($A^-$) with the columns associated to indices in $W_B$, and $A_{\overline W_{B}}^+$ ($A_{\overline W_{B}}^-$)  
the submatrix with the columns associated to those in $\overline W_{B}$. Similarly, submatrices $C_{W_{L}}$ and $C_{\overline W_{L}}$ are 
made up of the columns of $C$ associated to, respectively, indices in $W_{L}$ and in $\overline W_{L}$.

At decomposition step $s$, 
given the current feasible solution $(A^{+,s},A^{-,s},\bm{\beta}^s,{C}^{s})$ and working sets $W_{B}$ and $W_{L}$, the {\it proximal point} modification of the decomposition subproblem is as follows:
\vspace{-10pt}
\begin{alignat}{2}
     \label{pr:decformulation}
     \min_{A_{W_{B}}^+,A_{W_{B}}^-,\bm{\beta},C_{W_{L}}} &  \quad  O(A_{W_{B}}^+,A_{W_{B}}^-,\bm{\beta},C_{W_{L}})+ \\  
     & \quad \frac{\psi}{2}\left[\sum_{t \in W_{B}}\left(\|A_{\cdot t}^+-A_{\cdot t}^{+,s}\|^2+\|A_{\cdot t}^- -A_{\cdot t}^{-,s}\|^2\right)\ +  \right. \nonumber
             \left. \sum_{t \in W_{L}}\|C_{\cdot t}-C_{\cdot t}^{s}\|^2  \right] \nonumber\\
    s.t. \quad & \quad  C_{\cdot t}^T \vec 1 
    =1  \quad t\in W_{L}  \nonumber \\
   & \quad C_{k\cdot} \vec 1 
   \geq  1   \quad k=1,\ldots,K \nonumber \\ 
   & \quad  \beta_j \geq a_{jt}^+ + a_{jt}^-  \quad j=1,\dots,p,
    \quad t=1,\dots,\tau_{B}\nonumber \\
    & \quad A^+,A^-\in [0,1]^{p \times | \tau_{B} | }, \;\; \bm{\beta} \in [0,1]^{p}, \;\; C \in [0,1]^{K \times | \tau_{L} | }. & \nonumber
\end{alignat}
where $\psi \geq 0$, 
$\vec 1$ is the vector of all ones and $O(A_{W_{B}}^+,A_{W_{B}}^-,\bm{\beta},C_{W_{L}})$ denotes the decomposition version of function \eqref{pr:formulation} in which only variables $A_{W_{B}}^+$, $A_{W_{B}}^-$, $\bm{\beta}$, and $C_{W_{L}}$ are optimized while the other ones are kept fixed at the current values ${A_{\overline W_{B}}^+}^{s}$, ${A_{\overline W_{B}}^-}^{s}$, and ${C_{\overline W_{L}}}^{s}$.\par  
In general, in the design of decomposition algorithms for constrained nonlinear programs a proximal point term is added to the objective function to ensure some asymptotic convergence properties (see \cite{grippo2000convergence}). 
However, in the considered ML context we are more interested in the classification accuracy of the trained model rather than in the asymptotic convergence toward local or global solutions of \eqref{pr:decformulation}. 
A further reason not to focus on convergence issues is that in ML an excessive computational effort in solving the optimization training problem may lead to overfitting phenomena. 
Hence, the proposed decomposition scheme aims at obtaining a sufficiently accurate classification model in a limited CPU-time, in a sort of ``early-stopping" setting. 
Nonetheless, as highlighted for instance in \cite{palagi2005convergence}, adding a proximal point term in a decomposition subproblem may also have a beneficial effect from a numerical point of view, by ``convexifying" the objective function. This is the rationale for including it into \eqref{pr:decformulation}. 
\par
After (approximately) solving \eqref{pr:decformulation} and obtaining a solution
        $({A_{W_{B}}^+}^*,{A_{W_{B}}^-}^*,\bm{\beta}^*,{C_{W_{L}}}^*),$
the current solution of the original problem is updated as 
        \begin{gather}\label{eq:update}
            (A^{+,s+1},A^{-,s+1},\bm{\beta}^{s+1},C^{s+1})= 
            (({A_{W_{B}}^+}^*,{A_{\overline W_{B}}^+}^{s}),({A_{W_{B}}^-}^*,{A_{\overline W_{B}}^-}^{s}),\bm{\beta}^*,({C_{W_{L}}}^*,{C_{\overline W_{L}}}^{s})).        
        \end{gather}

The general decomposition scheme, referred to as NB-DEC (Node Based Decomposition), is shown in Algorithm \ref{algo:NB-DEC}.

        \begin{algorithm}\caption{NB-DEC}
           \label{algo:NB-DEC} \begin{algorithmic}
                \State set $s=0$, $\psi \geq 0$ and given $(A^{+,0},A^{-,0},\bm{\beta}^{0},C^{0})$ feasible for \eqref{pr:formulation} 
                \While{{\it stopping criterion}} 
 \vspace{-10pt}
                  \begin{enumerate}
                       \item select $W_{B}$ and $W_{L}$ as in \eqref{eq:BL} 
                       \item determine $({A_{W_{B}}^+}^*,{A_{W_{B}}^-}^*,\bm{\beta}^*,{C_{W_{L}}}^*)$ by solving (approximately) subproblem \eqref{pr:decformulation} 
                       \item set $s=s+1$ and $(A^{+,s+1},A^{-,s+1},\bm{\beta}^{s+1},C^{s+1})$ as in \eqref{eq:update}
                   \end{enumerate}
                   \vspace{-15pt}
                \EndWhile
                
            \Return $(A^{+,s},A^{-,s},\bm{\beta}^{s},C^{s})$ 
            \end{algorithmic}
        \end{algorithm}

In the NB-DEC initialization phase, a non-negative value is selected for the proximal point coefficient $\psi$ 
and a feasible starting solution $(A^{+,0},A^{-,0},\bm{\beta}^{0},C^{0})$ is provided. The main loop, which consists of three steps, is iterated until a certain stopping condition is met.
In the first step, the working set selection is performed. In particular, the indices associated to the branch and leaf nodes to be added to, respectively, $W_{B}$ and $W_{L}$ are selected. 
In the second step, the subproblem \eqref{pr:decformulation} is (approximately) solved to 
obtain the partial solution $({A_{W_{B}}^+}^*,{A_{W_{B}}^-}^*,\bm{\beta}^*,{C_{W_{L}}}^*)$. The latter is used in the third step to update the current solution. At the end of the main loop the current solution $(A^{+,s},A^{-,s},\bm{\beta}^{s},C^{s})$ is returned to build the classification tree.\par
        
The NB-DEC scheme is very general and may encompass 
a variety of 
different versions. Indeed, the stopping criterion, the working set selection rule 
and the algorithm to solve the subproblem \eqref{pr:decformulation} are not specified. Concerning the stopping criterion, different choices are possible. For instance, it can be related to the satisfaction of the optimality conditions with respect to the original problem \eqref{pr:formulation}, to the accuracy 
of the classification model on a certain validation set, or to a maximum budget 
in terms of iterations or of CPU-time. \par 
The asymptotic convergence of the algorithm strongly depends on the working set selection rule and the way subproblem \eqref{pr:decformulation} is solved (see \cite{grippo2000convergence} for convergence conditions). As previously pointed out, the focus here is to produce a sufficiently accurate model in short CPU-time. This can be generally achieved by reducing as much as possible the regularized loss function within a limited budget of CPU-time or iterations. From this point of view, a more suitable requirement might be the monotonic decrease of the loss function, i.e. 
        \begin{equation}\label{eq:decrease}
            O(A^{+,s+1},A^{-,s+1},\bm{\beta}^{s+1},C^{s+1})\leq O(A^{+,s},A^{-,s},\bm{\beta}^{s},C^{s}).
        \end{equation}
Since $\psi \geq 0$, 
it is easy to see that \eqref{eq:decrease} is ensured by applying any descent algorithm with any degree of precision in the solution of \eqref{pr:decformulation} at the second step.
\subsection{Comments on theoretically convergent versions}\label{subsec:convergent}
Although in the considered framework the asymptotic convergence is not the main concern, it is easy to derive versions of 
	NB-DEC satisfying the global convergence property stated in \cite{grippo2000convergence}. Indeed, let us consider a 
	NB-DEC version, referred to as C-NB-DEC (Convergent Node Based Decompositon), in which the working set selection at instruction 2. of Algorithm \ref{algo:NB-DEC} is performed as an alternation of the following two choices:
\begin{enumerate}
    \item[(i)] $W_{\cal B}=\{1,\dots, | \tau_{\cal B}|\}, \quad W_{\cal L}=\{\emptyset \}$ (full branch nodes and empty leaf nodes working set),
    \item[(ii)] $W_{\cal B}=\{\emptyset\}, \quad W_{\cal L}=\{1,\dots, | \tau_{\cal L}| \}$ (empty branch nodes and full leaf nodes working set).
\end{enumerate}
Then, the feasible set would consist in the Cartesian product of closed convex sets with respect to the variables' blocks involved in each of the two types of working sets. Since the feasible set of every decomposition subproblem is compact and the objective function is continuous, by the Weierstrass Theorem each subproblem admits an optimal solution, so it is well defined as stated in \cite{grippo2000convergence}. Moreover, since the sequence $\{(A^{+,s},A^{-,s},\bm{\beta}^{s},C^{s})\}$ produced by 
C-NB-DEC is defined over a compact feasible set, it admits limit points. Hence, considering also the presence of the proximal point term, from Proposition 7 of \cite{grippo2000convergence} every limit point of the sequence produced by 
C-NB-DEC is a critical point for \eqref{pr:decformulation} (a feasible point is critical if no feasible descent directions exist at that point).\par

Notice also that if the indices in $W_{\cal B}$ at step (i) are divided into any partition and step (i) is splitted into a sequence of internal decomposition steps 
based on the considered partition, then the above-mentioned convergence property still holds (provided that the same partition is adopted at every step (i)).

         
\subsection{S-NB-DEC: an efficient practical version}  
\label{subsec:practical_dec}

Despite its asymptotic convergence property, 
C-NB-DEC algorithm showed in preliminary experiments (not reported here for brevity) to be not that efficient as it is not suited to fully exploit the decomposition of the general scheme and the intrinsic structure of problem \eqref{pr:decformulation}. \par

For this reason, here we present an efficient practical version of NB-DEC, referred to as S-NB-DEC (Single branch Node Based Decomposition), and compare it to the 
method without decomposition in order to test the benefits of the decomposition approach. Even though S-NB-DEC is a heuristic, it adopts an ``intense" branch nodes decomposition that makes it more efficient than the not decomposed version and the aforementioned convergent C-NB-DEC.\par 
S-NB-DEC is 
obtained from NB-DEC by specifying the stopping criterion, the working set selection rule and the subproblem solver.
In particular, at each decomposition step $s$, only a single index associated to a random branch node is inserted in $W_{B}$, while all indices associated to the leaf nodes are inserted in $W_{L}$. Each branch node is randomly selected only one time per {\it macro-iteration}, i.e., a sequence of decomposition steps in which all the branch nodes have been selected one time in $W_{B}$.\par

         
As stopping criterion a maximum number of macro-iterations is adopted. An 
alternative criterion could be related to the satisfaction of the Karush-Khun-Tucker conditions for problem \eqref{pr:formulation}, but as previously mentioned the former is more suitable in case of limited training time.
\par 

The S-NB-DEC method, summarized in Algorithm \ref{algo:VCdim}, mainly consists of two nested loops. 
In the internal one, multiple decomposition steps are performed until all the branch nodes are randomly selected from the set List. The partial working set $W_{B}$ is constructed (instruction 6) on the basis of the random selection operated at instruction 5. It is worth mentioning that $W_{L}$ is always made up of indices of all leaf nodes for stability reasons, as changing the variables of a single branch node 
affects the optimality of the variables associated to all leaf nodes. However, this does not represent a significant limitation as the number of leaf nodes variables is not that large for most practical problems.\par 
S-NB-DEC can optionally be started with an initialization step in which all variables are inserted in the working set (no actual decomposition is performed) and a limited number of iterations ($init\_iter$) of an NLP solver is applied to formulation \eqref{pr:decformulation} (which in this case coincides with \eqref{pr:formulation}). The solution 
obtained at the end of this phase (denoted as $(A^{+,init\_iter},A^{-,init\_iter},\bm{\beta}^{init\_iter},C^{init\_iter})$), will be used as initial solution for the subsequent decomposition phase. In some cases, the Initialization may improve the stability of the method by providing the decomposition algorithm with more promising starting solutions, 
as the latter are obtained by using all variables' information. However, the Initialization may be out of reach for very large instances. 
Notice that, if no Initialization is applied, the starting solution $(A^{+,0},A^{-,0},\bm{\beta}^{0},C^{0})$ must be provided otherwise. In such cases, whenever the number of leaf nodes is larger or equal to the number of classes, 
an initial feasible 
solution can be easily obtained by setting to zero all variables $a_{jt}$ and $\beta_j$ and setting $c_{kt}=1/K$ with $k=1,\dots,K$ and $t=1,\dots,\tau_{L}$.
 \begin{algorithm}[H]\caption{S-NB-DEC}
        \label{algo:VCdim}
            \begin{algorithmic}[1]
            \Statex {} 
            
           \State set $s=0$, $\psi \geq 0$, $W_{B}=\{1,2,\dots, | \tau_{B} | \}$,  $W_{L}=\{1,2,\dots, | \tau_{L} | \}$, $max\_iter>0$ \If{Initialization == True} \Comment{{\tt Initialization} (optional)}
	    \State  set  $init\_iter>0$ 
               \State apply NLP solver for $init\_iter$
               iterations to \eqref{pr:decformulation}
             \State set  $(A^{+,0},A^{-,0},\bm{\beta}^{0},C^{0}) = (A^{+,init\_iter},A^{-,init\_iter},\bm{\beta}^{init\_iter},C^{init\_iter})$
            \Else
            \State set $(A^{+,0},A^{-,0},\bm{\beta}^{0},C^{0})$ provided as input
          \EndIf
               
                
                \While{$s<max\_iter$ } \Comment{{\tt Decomposition}}
                   \State set List=$\{1,2,\dots, | \tau_{B} | \}$
                   \While{List$\neq \emptyset$}
                        \State select $\bar \imath$ randomly from List and set List=List$\setminus 
                        \{ \bar \imath \}$
                        
                        \State  set $W_{B}=\{\bar \imath\}$,\;\;\;\;  $W_{L}=\{1,2,\dots, | \tau_{L} | \}$
                        \State determine $({A_{W_{B}}^+}^*,{A_{W_{B}}^-}^*,\bm{\beta}^*,{C_{W_{L}}}^*)$ by applying NLP solver to  \eqref{pr:decformulation} 
                        \State set $(A^{+,s+1},A^{-,s+1},\bm{\beta}^{s+1},C^{s+1})$ as in \eqref{eq:update}
                    \EndWhile
                    \State set $s=s+1$  
                \EndWhile
                
            \hspace{-25pt}\Return $(A^{+,s},A^{-,s},\bm{\beta}^{s},C^{s})$
            \end{algorithmic}
        \end{algorithm}

\subsection{Numerical results} 
In this section, the proposed decomposition algorithm S-NB-DEC is compared with the
not decomposition strategy, 
referred to as  not-DEC, for the optimization of the MRCT $L_{0}^{glob}$ formulation. In order to 
assess the benefits of the decomposition on 
larger datasets, 
we consider the two largest datasets of Section \ref{Datasets and experiments} (Thyroid and Car)
        as well as three additional
        large ones (Splice, Segment and Dna), see Table \ref{tab:newdata}. 
        \begin{table}[H]
            \caption{Datasets for testing decomposition}
            \label{tab:newdata}
            \centering
           \scalebox{0.7}{%
            \begin{tabular}{lllllll}

            \hline
            Dataset                                                                 & Abbreviation        & N    & p   & K &  Class distribution     & Proximal ($\psi$)       \\ \hline
            Car-evaluation &
              Car &
              1728 &
              15 &
              4 &
              70\%-22\%-4\%-4\% & 1.25$\text{e}^{-4}$ \\
            
            Thyroid-disease-ann-thyroid &
              Thyroid &
              3771 &
              21 &
              3 &
              92.5\%-5\%-2.5\% &  1.25$\text{e}^{-6}$\\

            Splice-junction Gene Sequences   & Splice    & 3190  & 60 & 3 & 51.9\%-24.1\%-24\%       & 2.5$\text{e}^{-7}$   \\
            Dna & Dna     & 3168 & 180 & 3 & 51.9\%-24.1\%-24.0\%         & 1.25$\text{e}^{-8}$ \\
            Image Segmentation & Segment          & 32310  & 19  & 7 & 14.3\%-14.3\%-14.3\%-14.3\%-14.3\%-14.3\%-14.3\% & 2.5$\text{e}^{-8}$ \\ \hline
            \end{tabular}
            }
        \end{table}
For S-NB-DEC,
the same IPOPT solver (as in Section \ref{sec:comp_results} for not-DEC) 
is used to solve subproblem \eqref{pr:decformulation}. 
Since we are not interested in accurately solving each subproblem, 
the maximum number of IPOPT iterations has been set to $40$, 
while the default value of $1e^{-8}$ has been adopted for the optimality tolerance. For the smaller datasets (Thyroid and Car), the Initialization step of S-NB-DEC has been enabled by running IPOPT on formulation \eqref{pr:formulation} for a limited number of internal interations (five), while for the larger datasets (Splice, Segment and Dna) a pure decomposition is applied without Initialization, as the latter would have been too computationally expensive.\par
S-NB-DEC is tested with and without a proximal point term. For all datasets an ``outer" 5-fold cross-validation has been
used by selecting randomly, for each fold, a fraction of $1/5$ of the samples 
for the testing set and keeping the remaining $4/5$ as training block. Then, a further ``inner" 5-fold cross-validation is applied to every training block to determine the best value of the proximal point parameter $\psi$. In particular, for each fold, the training block is splitted into a random fraction of $1/5$ of the samples 
used as validation, and the remaining $4/5$ is actually used for 
training. The training of each one of the 5 inner folds is performed for each one of the four values of $\psi$ in $\{ 1.25e^{-3},1.25e^{-4},1.25e^{-5},1.25e^{-6} \}$ and the accuracy of the resulting models are then evaluated on the corresponding validation sets. 
To cope with the nonconvexity of the training problem, for each combination of the $\psi$ value and inner fold, the training is repeated 10 times from 10 different random initial 
solutions (the IPOPT solver fixes automatically any infeasibility of the random initial solutions). The accuracy on the validation set associated to each $\psi$ value is averaged first over the 10 runs and then 
over the 5 outer folds. The $\psi$ value obtaining the overall best performance, say 
$\hat{\psi}$ is selected for the final outer cross-validation, in which 10 runs of training are performed again from 10 different starting solutions 
for each outer training fold (including both the inner training and validation sets). The final accuracy is obtained by averaging the accuracy on the testing sets over 10 runs and over the 5 outer folds. When the proximal point term is not included in the 
formulation, only the outer cross-validation is performed. For Splice, Segment and Dna, the inner cross-validation used to obtain the best $\psi$ value has not been applied due to 
high training times  
and $\hat{\psi}$ has been determined by a 
simple
rule of thumb based on 
their size. 
\par

Concerning $\lambda_{0}^G$, the same values of Section \ref{Datasets and experiments} have been used for Car and Thyroid ($0.25$ and $0.5$ respectively). For Splice, Segment and Dna also the 
sparsity parameters have been 
empirically derived from their size 
($0.5$ for all three).\par
The adopted IPOPT options for not-DEC are the default ones ($1e^{-8}$ as optimality tolerance and $3000$ as maximum number of iterations). 
Preliminary experiments showed that reducing the precision of the optimality tolerance or the maximum number of iterations for the 
method without decomposition did not 
yield, in general, accurate enough models (see below for the model quality after five iterations of not-DEC on the Car and Thyroid datasets).\par

\begin{figure}[H]

   \begin{minipage}{0.45\textwidth}
     \centering
     \textbf{without proximal term}\par\medskip
     \includegraphics[width=1\linewidth]{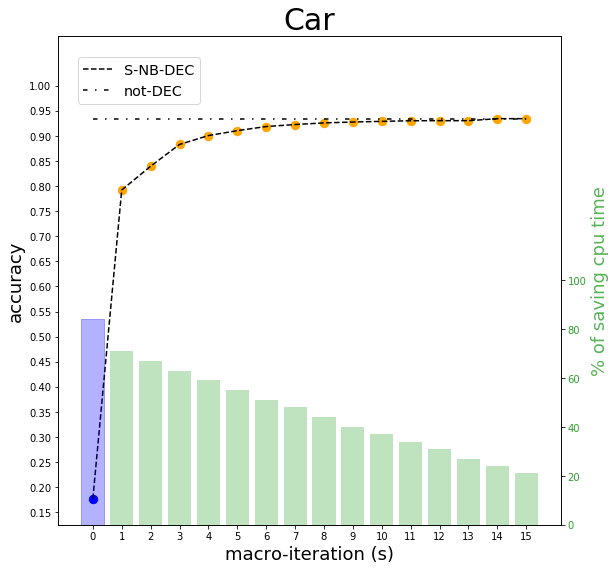}
   \end{minipage}
   \hfill
   \begin{minipage}{0.45\textwidth}
     \centering
     \textbf{with proximal term}\par\medskip
     \includegraphics[width=1\linewidth]{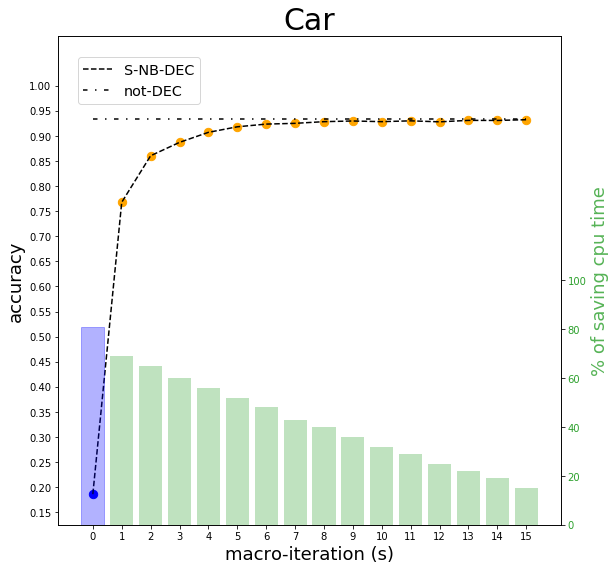}

   \end{minipage}
   
   \begin{minipage}{0.45\textwidth}
     \centering
     \includegraphics[width=1\linewidth]{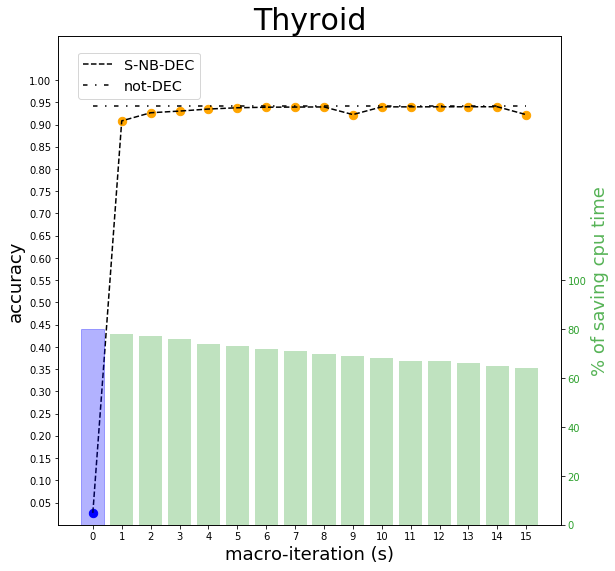}
      \label{Fig:Data1}
   \end{minipage}\hfill
   \begin{minipage}{0.45\textwidth}
     \centering
     \includegraphics[width=1\linewidth]{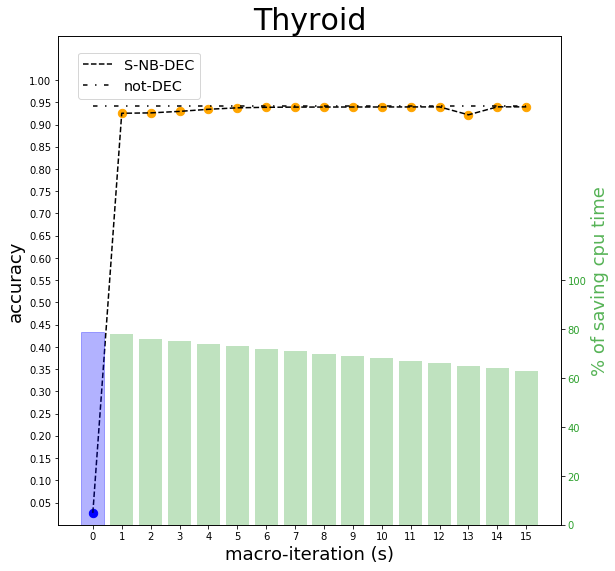}

   \end{minipage}
   
   \caption[]{Results of the  MRCT $L_{0}^{glob}$ formulation using S-NB-DEC and not-DEC for the Car and Thyroid datasets. The blue bar represents the Initialization step consisting of 5 interations of not-DEC (namely without decomposition).}
   \label{fig:dec-plots1}
\end{figure}

\begin{figure}[H]
    
    \begin{minipage}{0.45\textwidth}
     \centering
    \textbf{without proximal term}\par\medskip
     \includegraphics[width=1\linewidth]{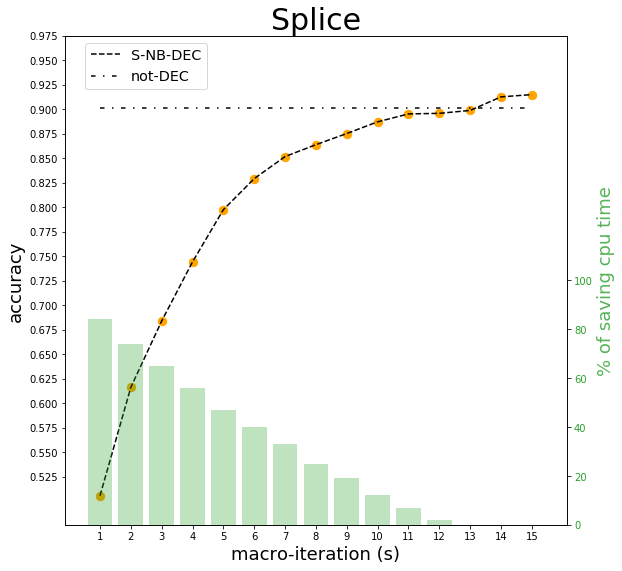}

   \end{minipage}
   \hfill
   \begin{minipage}{0.45\textwidth}
     \centering
     \textbf{with proximal term}\par\medskip
     \includegraphics[width=1\linewidth]{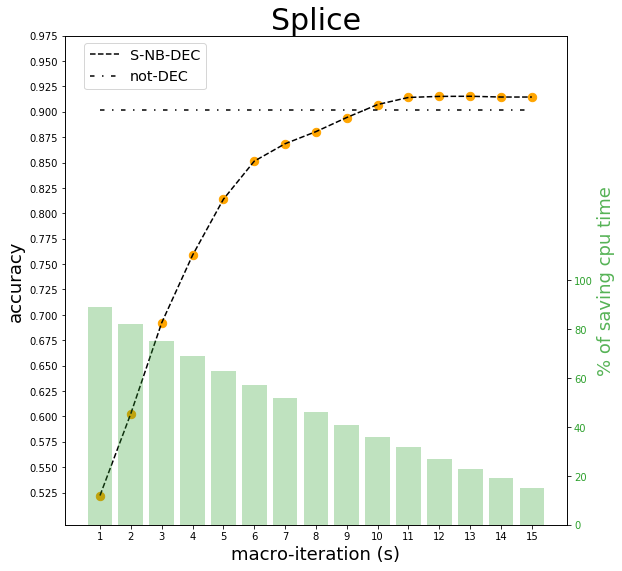}
     
   \end{minipage}
     \begin{minipage}{0.45\textwidth}
     \centering
     \includegraphics[width=1\linewidth]{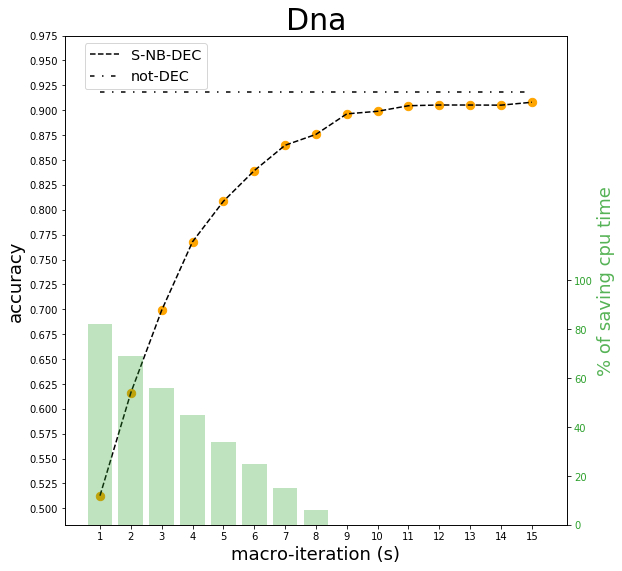}
     
   \end{minipage}\hfill
   \begin{minipage}{0.45\textwidth}
     \centering
     \includegraphics[width=1\linewidth]{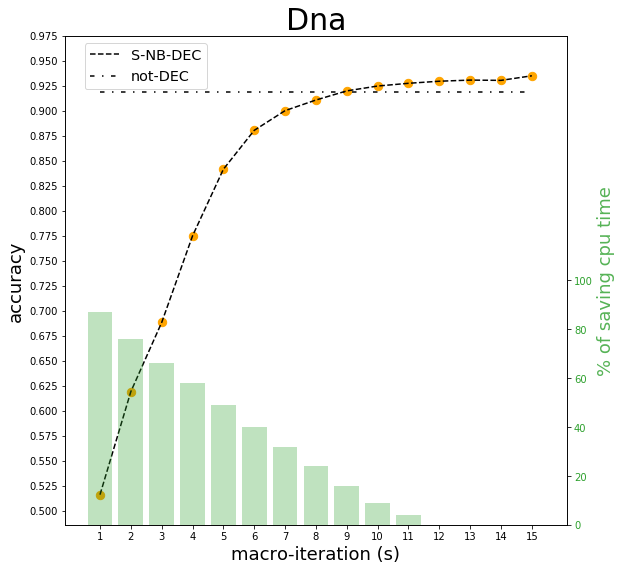}
     
   \end{minipage}
   \begin{minipage}{0.45\textwidth}
     \centering
     \includegraphics[width=1\linewidth]{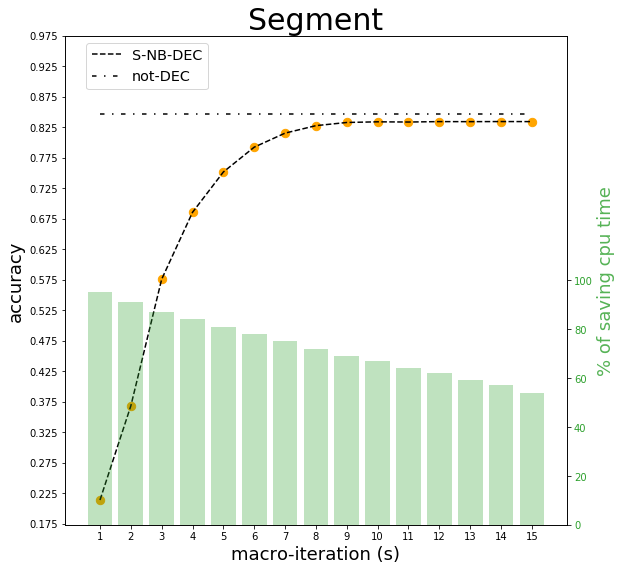}
     
   \end{minipage}\hfill
   \begin{minipage}{0.45\textwidth}
     \centering
     \includegraphics[width=1\linewidth]{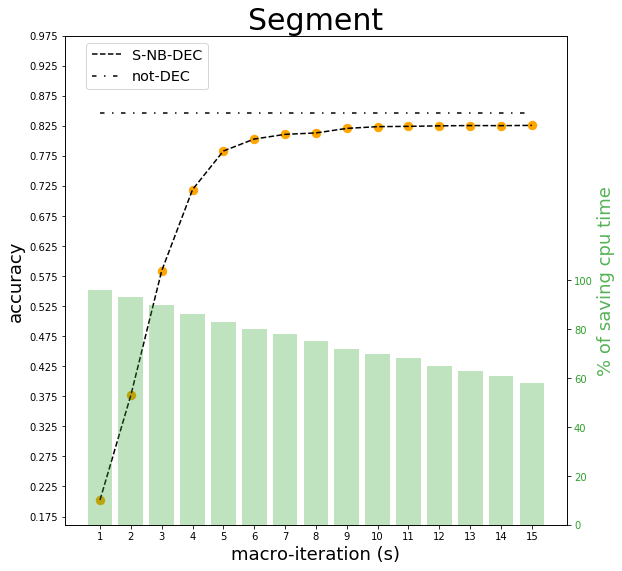}
     
   \end{minipage}
   
   \caption[]{Results of the  MRCT $L_{0}^{glob}$ formulation using S-NB-DEC and not-DEC for the Splice, Dna and Segment datasets.}
   \label{fig:dec-plots2}
\end{figure}

S-NB-DEC is compared with not-DEC in terms of testing accuracy and CPU-time needed to train the 
trees. 
Also the training times and the testing accuracy are averaged over the 10 runs and the 5 outer folds. 
The synoptic plots of Figures \ref{fig:dec-plots1} and \ref{fig:dec-plots2} depict the results of the numerical comparison between S-NB-DEC and not-DEC, both in terms of testing accuracy and CPU-time. In particular, the x-axis represents the macro-iterations of the decomposition algorithm, the left y-axis represents the testing accuracy and the right y-axis (highlighted in green) represents the percentage of CPU-time saving obtained with the decomposition algorithm. In correspondence of each macro-iteration, the accuracy level obtained 
with the decomposition algorithm (dashed profile) is marked with a yellow circle and the percentage of CPU-time saving with respect to not-DEC is depicted as a vertical green bar. The horizontal dash-dotted line represents the accuracy level of not-DEC. If the Initialization step is applied, the first blue circle and bar refer, respectively, to the testing accuracy and to the CPU-time measured at the end of the Initialization.\par 
Firstly let us consider the two smaller datasets.
Concerning Car, both the versions of S-NB-DEC with and without the proximal point term
achieve the same accuracy of not-DEC with a CPU-time saving greater than $30\%$ (after $9$ and $11$ macro-iterations respectively). As to Thyroid, both versions achieve the same accuracy of not-DEC at macro-iteration $6$ with a CPU-time saving greater than $60\%$. Notice that for both datasets, even if the solution 
obtained at the end of the Initialization step has a poor accuracy, after only a few decomposition macro-iterations S-NB-DEC is able to approximately achieve the same accuracy reached by not-DEC (especially for Thyroid in which just one decomposition step is enough to obtain a very good model). From a CPU-time point of view, running the $5$ IPOPT iterations of the Initialization step is much more time consuming than a single decomposition macro-iteration.
However, preliminary experiments (not reported here for brevity) showed that the Initialization step, whenever computationally viable, helps in speeding up and stably driving the subsequent decomposition steps towards good quality solutions.

\par 
Let us consider the three larger datasets, for which the Initialization has been disabled.     
As to the Splice dataset, 
S-NB-DEC without the proximal point term 
yields trees with a better accuracy than not-DEC within approximately the same CPU-time, 
while the proximal point version achieves the same 
accuracy as not-DEC at macro-iteration 11 ($30\%$ CPU-time saving) and then improves it from macro-iterations $12$ to $15$ (CPU-time savings are comprised between $30\%$ and $15\%$). \par 
        
Concerning the Segment dataset, the one with the largest number of samples,  
S-NB-DEC without proximal point term 
provides trees with
a testing accuracy lower than 
that of not-DEC by 
about $1\%$, but with a CPU-time saving of almost $70\%$ (at macro-iteration 9), while the proximal point version obtains similar results but 
with a slightly lower accuracy
than that without proximal term.\par

For the
Dna dataset, the one with largest number of features, the S-NB-DEC 
yileds trees with
an accuracy loss of about $2.5\%$ with respect to not-DEC, without any significant CPU-time saving, while the proximal point version achieves the same accuracy of not-DEC with a CPU-time saving of almost $20\%$ (macro-iteration $9$) and it is also able to slightly improve the accuracy in correspondence of CPU-time savings between $10\%$ and $5\%$ (macro-iterations $10$ and $11$) or for slightly larger CPU-times.\par 

To summarize, the above numerical  
results 
indicate that the S-NB-DEC decomposition approach allows to significantly reduce the computational time needed to solve the training problem \eqref{pr:formulation}, without compromising too much the classification trees accuracy.
In certain cases, the decomposition 
even yields improved testing accuracy (see Splice and Dna). 
Whenever applicable, the Initialization step may facilitate 
faster progress towards a good 
quality solution, although for larger datasets the pure decomposition is 
able to achieve promising results both in terms of accuracy and CPU-time savings.\par
As expected, the presence of the proximal point term is often helpful in speeding up the training process. 
\section{Concluding remarks}
\label{sec:concluding}

We have investigated the interesting nonlinear optimization formulation proposed in \cite{blanquero2018optimal,blanquero2020sparsity} for training (sparse) MRCTs along three directions.
First, we presented alternative methods to sparsify MRCTs based on concave approximations of the $l_0$ ``norm" and we compared them with the original $l_1$ and $l_{\infty}$ regularizations. 
Second, we derived lower and upper bounds on the VC dimension of MRCTs. Third, we proposed a general proximal point decomposition scheme to tackle larger datasets and we described an efficient version of the method. 

The results reported for 24 datasets indicate that the alternative sparsification method based on approximate $l_0$ regularization compares favourably with the original approach and leads to more compact MRCTs. Moreover, the decomposition method yields promising results in terms of speed up and of testing accuracy on five larger datasets. Note that achieving a significant speed up in the training of MRCTs while maintaining comparable accuracy allows to widen the range of applicability of such ML models. This may also constitute a step toward the combination of such MRCTs, with other models or à la ensemble, in an attempt to further improve 
accuracy.

Future work includes investigating different working set selection strategies for the decomposition and extending these decomposition methods to deal with additional side constraints such as cost-sensitivity and fairness as outlined in \cite{carrizosa2021mathematical}.



        

\bibliography{mybibfile}

\newpage

\hypertarget{appendix}{}

\section*{Appendix: Heatmaps}

\begin{figure}[H]
\centering
\includegraphics[width=\textwidth]{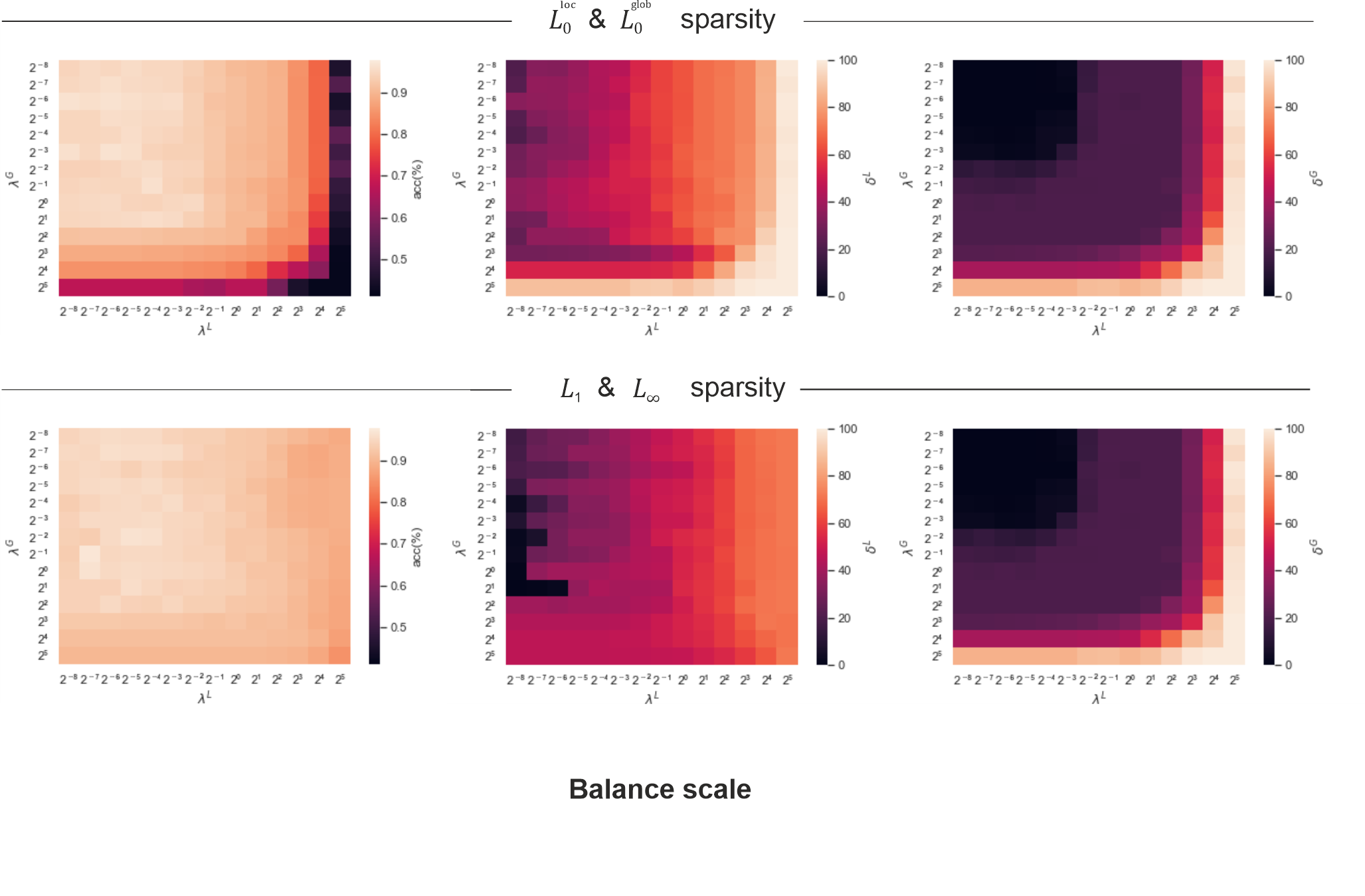}
\end{figure}

\begin{figure}
\centering

\includegraphics[width=\textwidth]{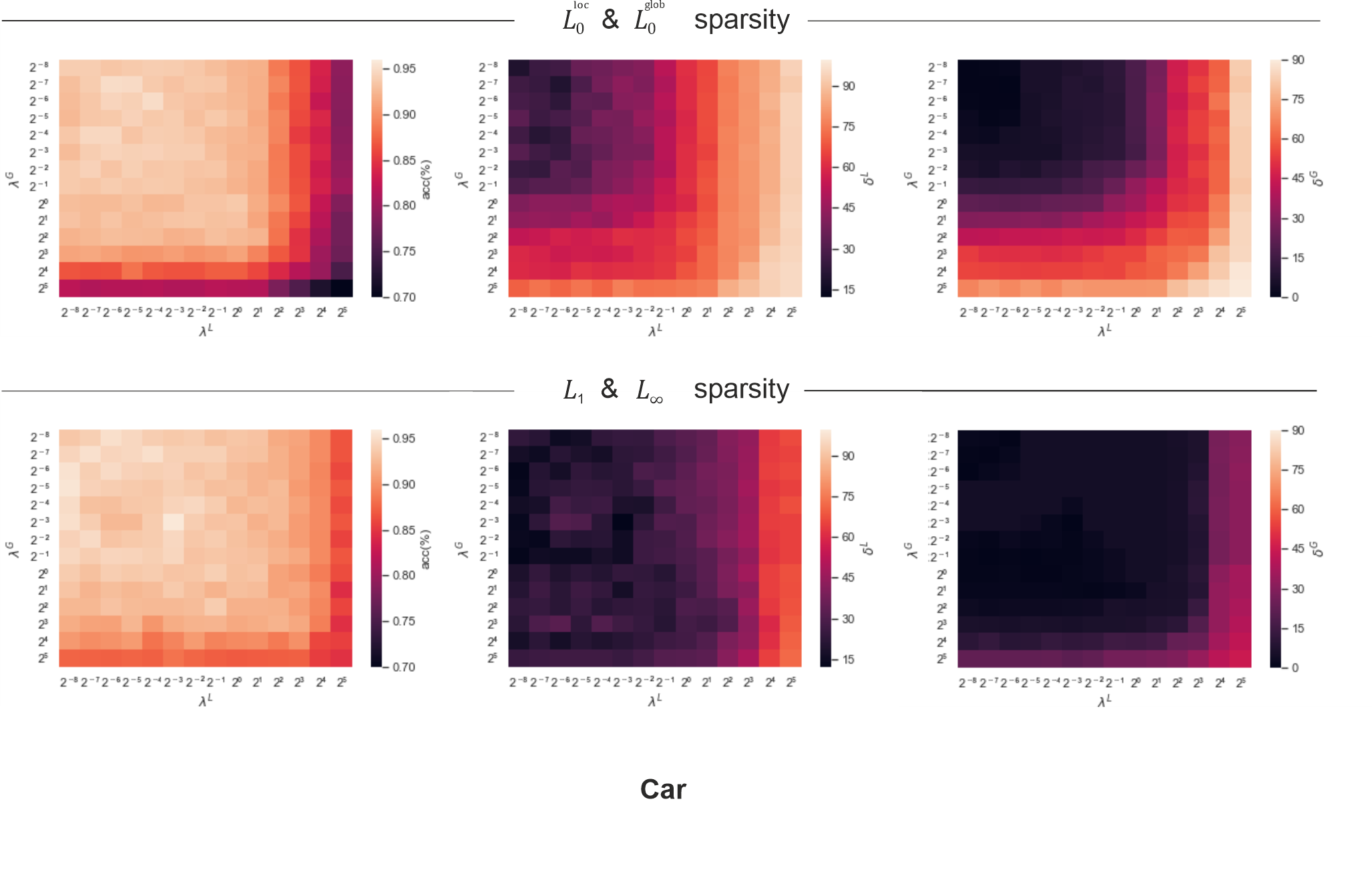}
\end{figure}

\begin{figure}
\centering

\includegraphics[width=\textwidth]{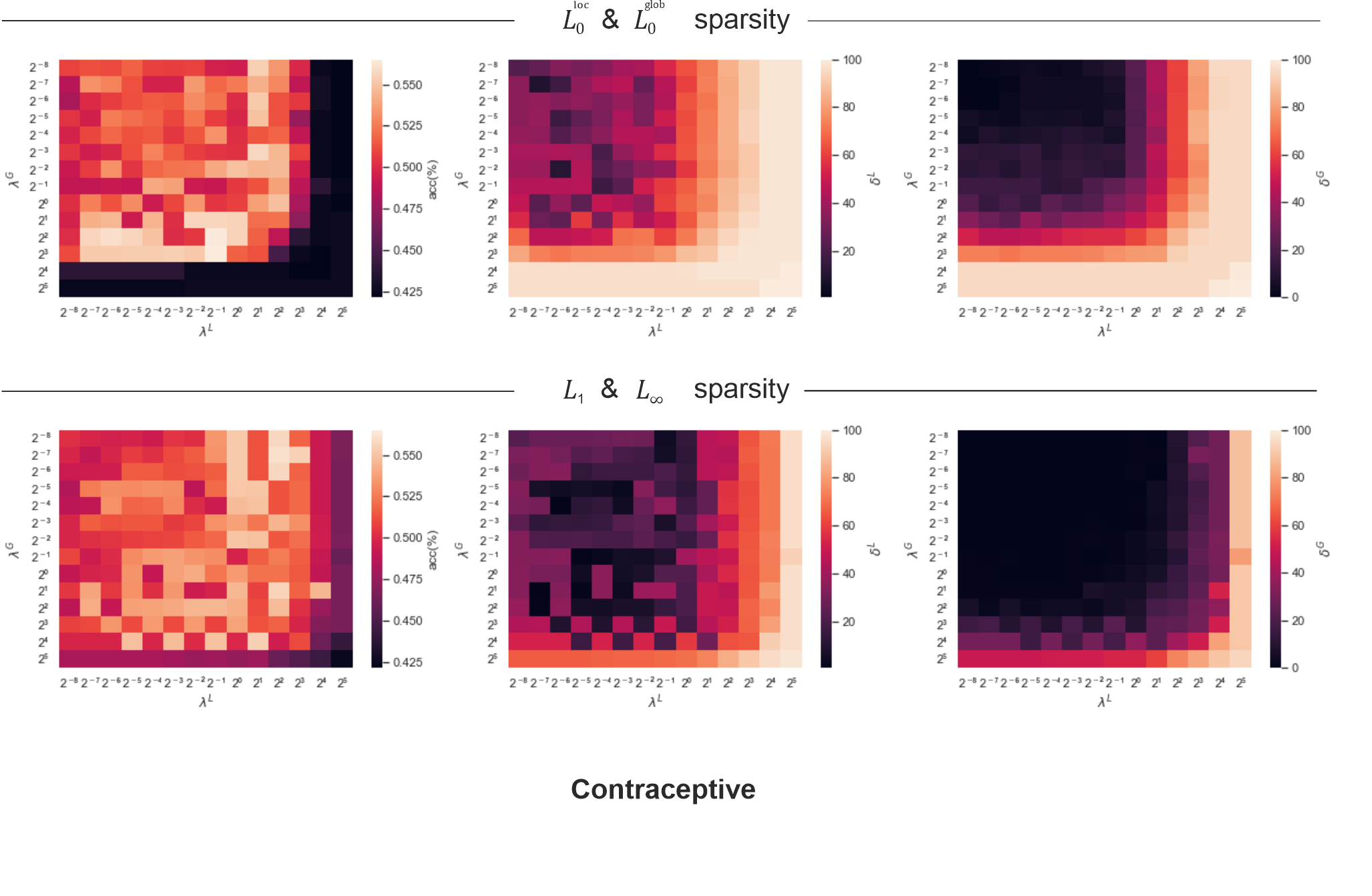}
\end{figure}

\begin{figure}
\centering

\includegraphics[width=\textwidth]{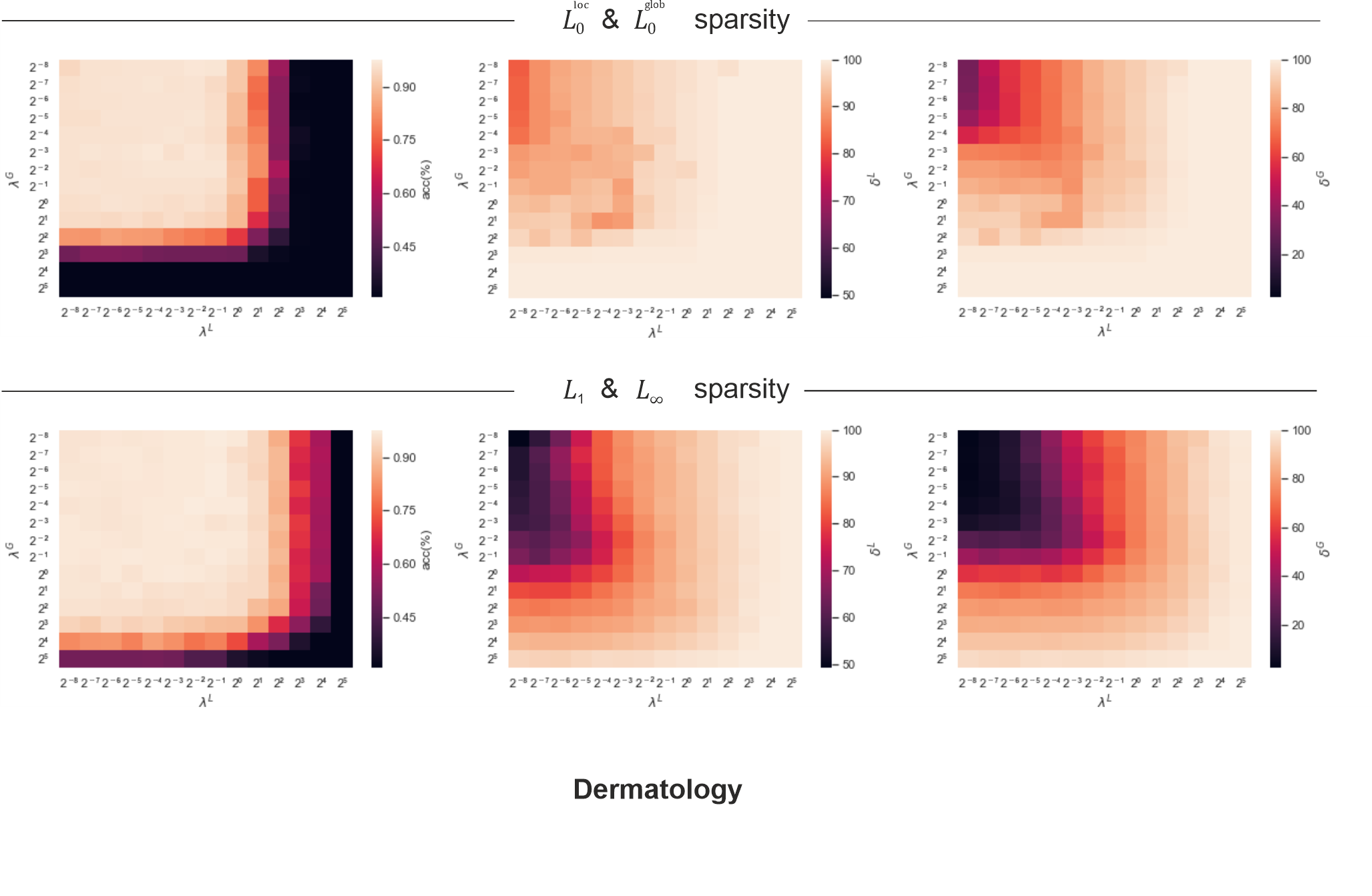}

\end{figure}
\begin{figure}
\centering

\includegraphics[width=\textwidth]{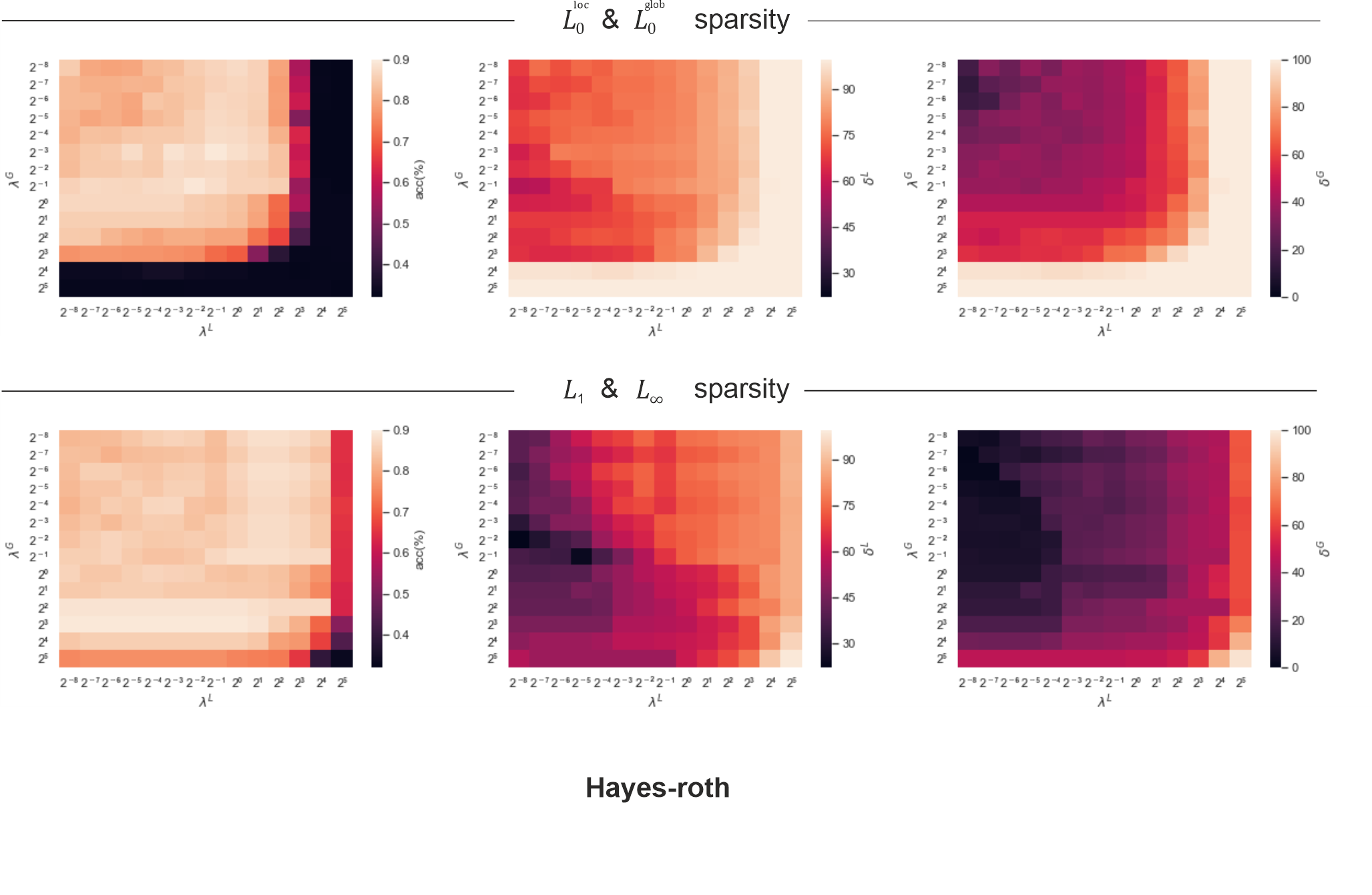}
\end{figure}

\begin{figure}
\centering

\includegraphics[width=\textwidth]{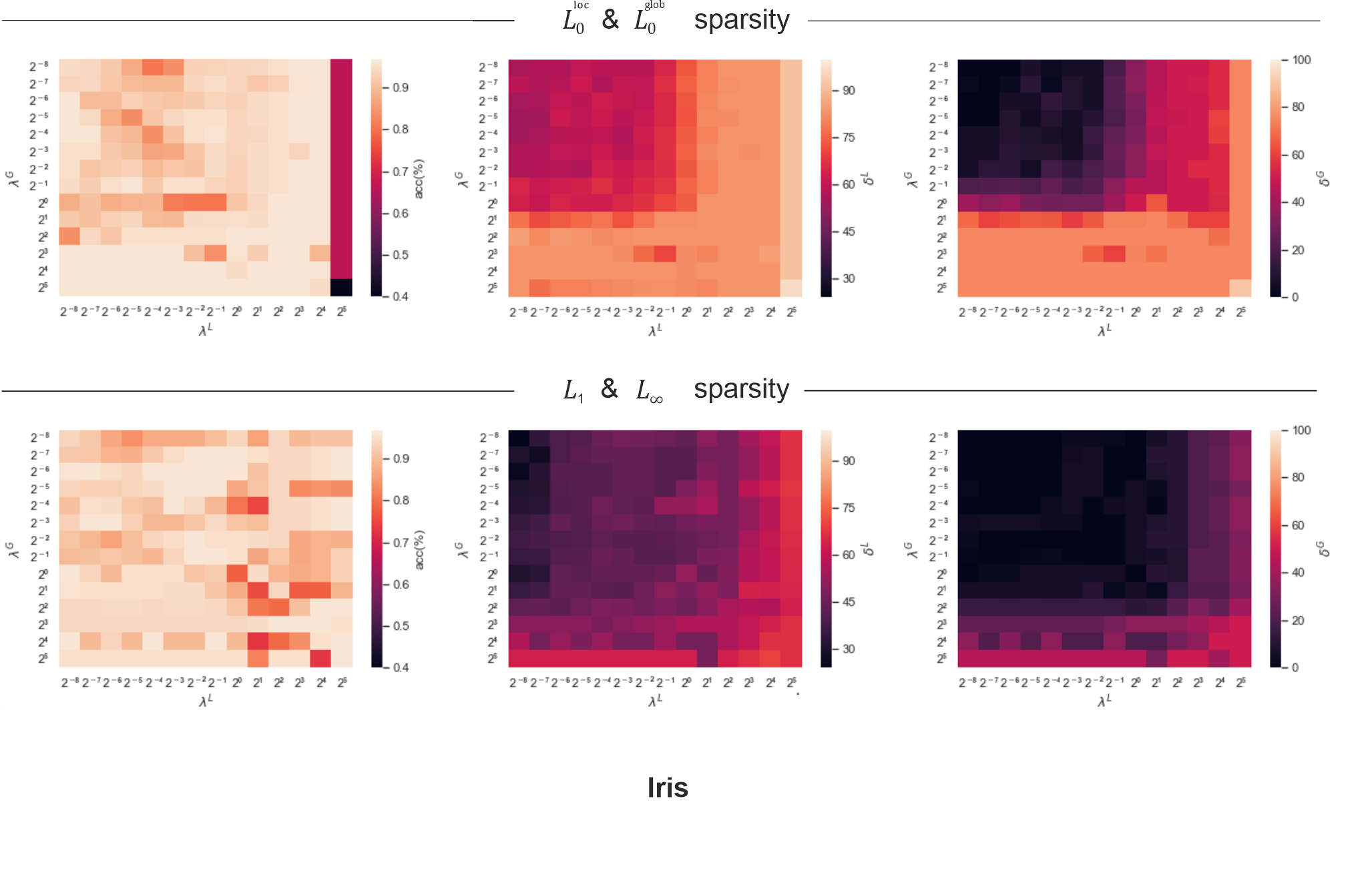}

\end{figure}

\begin{figure}
\centering
\includegraphics[width=\textwidth]{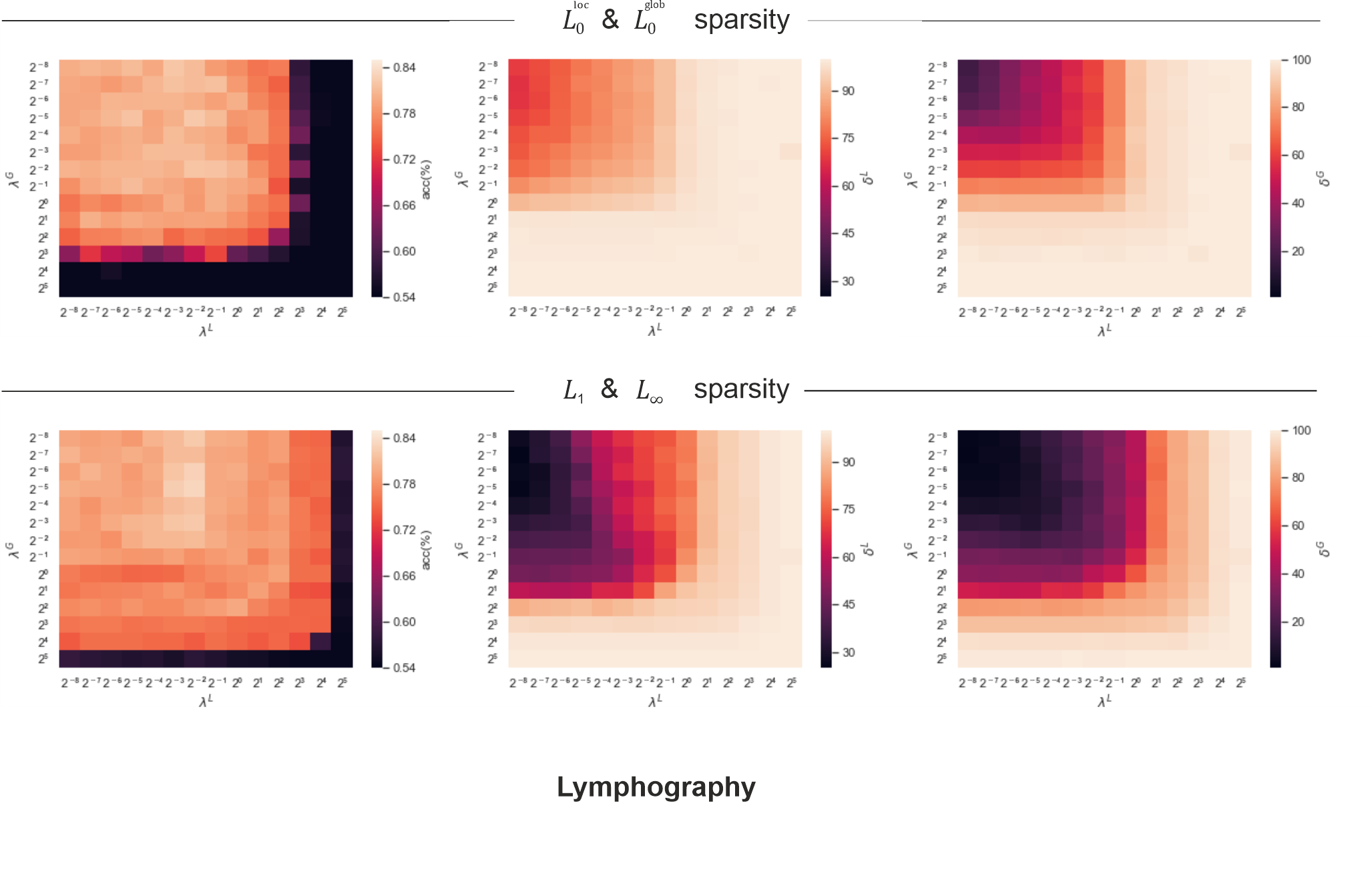}

\end{figure}

\begin{figure}
\centering

\includegraphics[width=\textwidth]{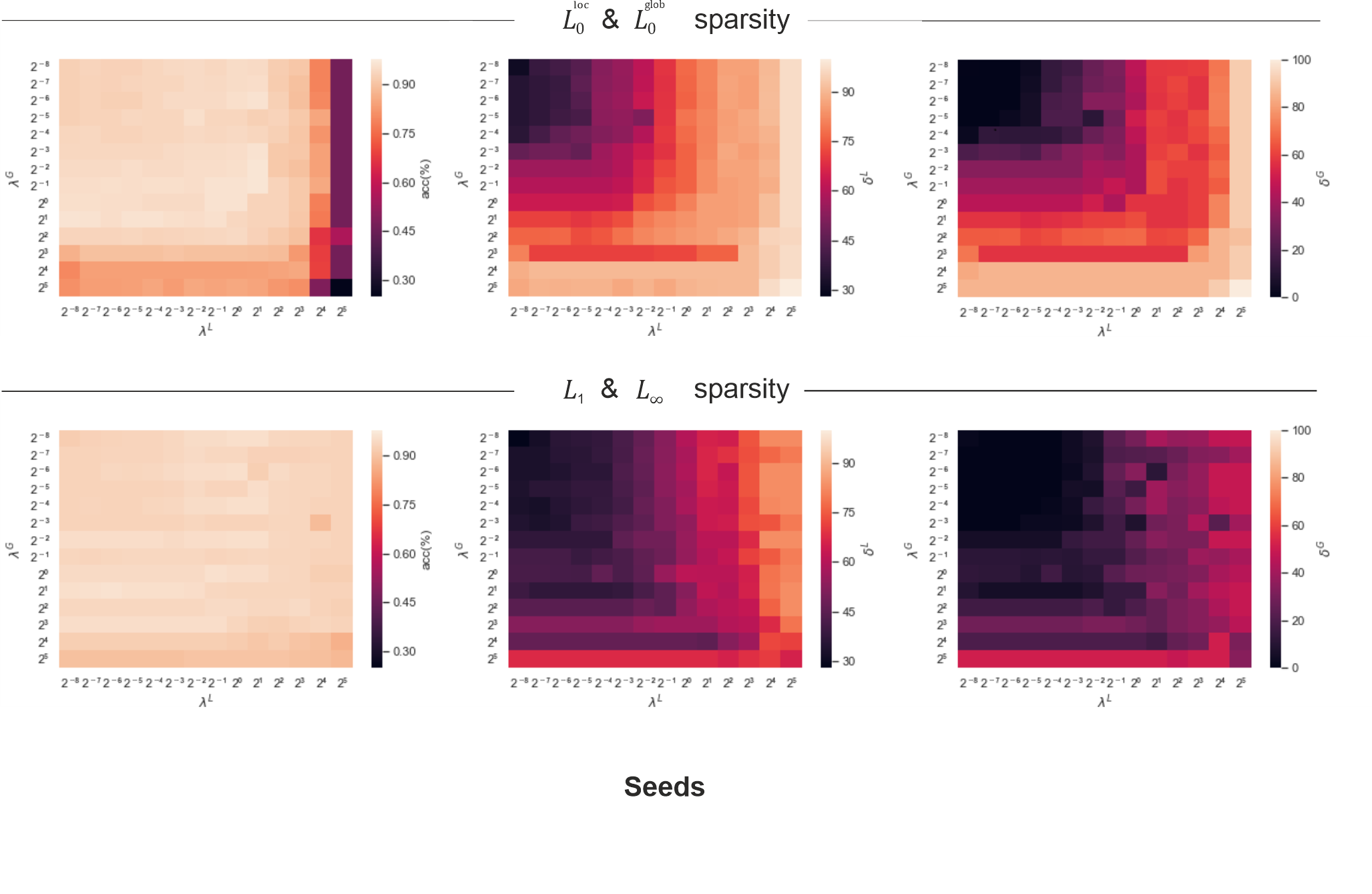}

\end{figure}
\begin{figure}
\centering

\includegraphics[width=\textwidth]{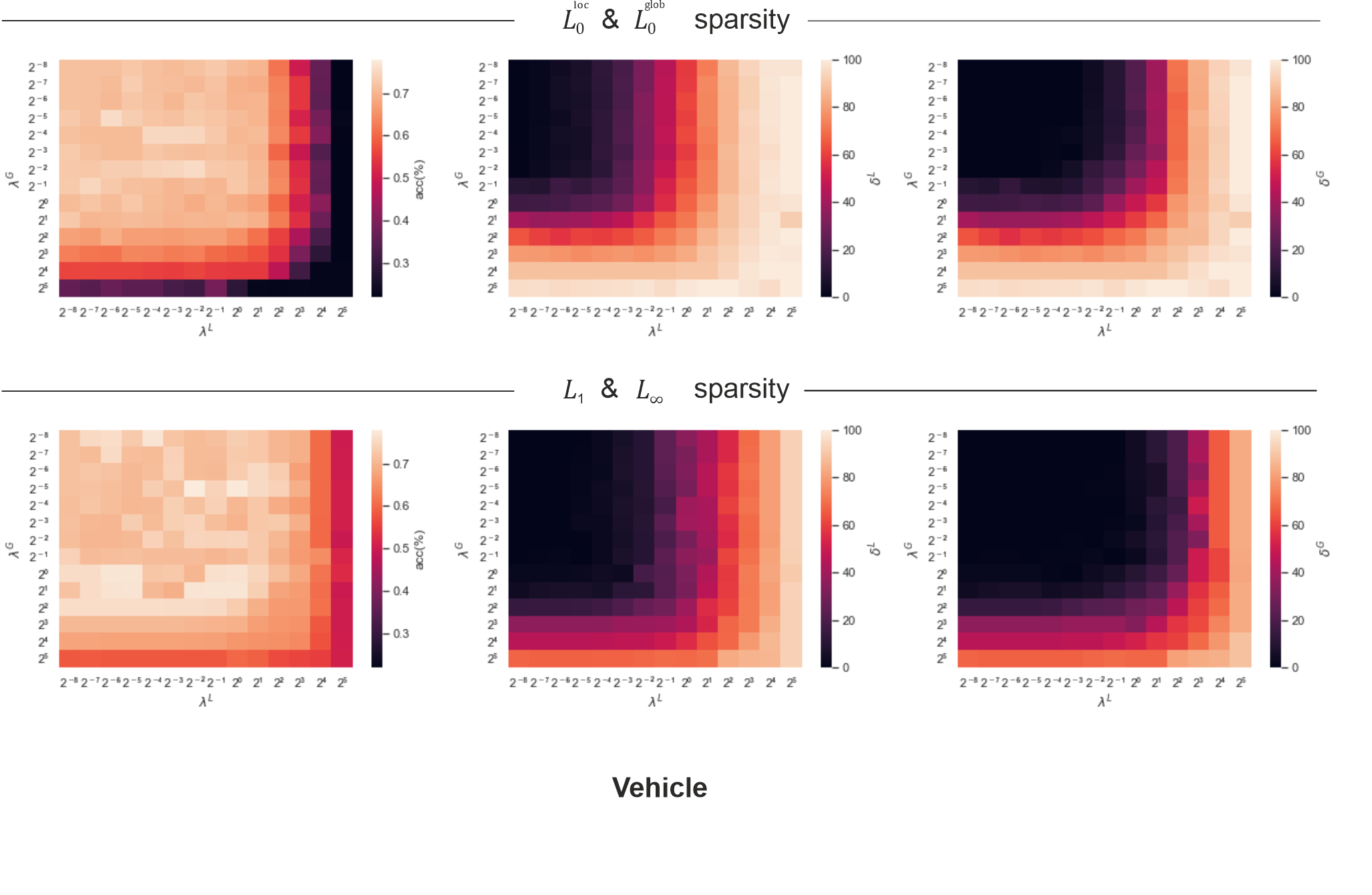}
\end{figure}

\begin{figure}
\centering

\includegraphics[width=\textwidth]{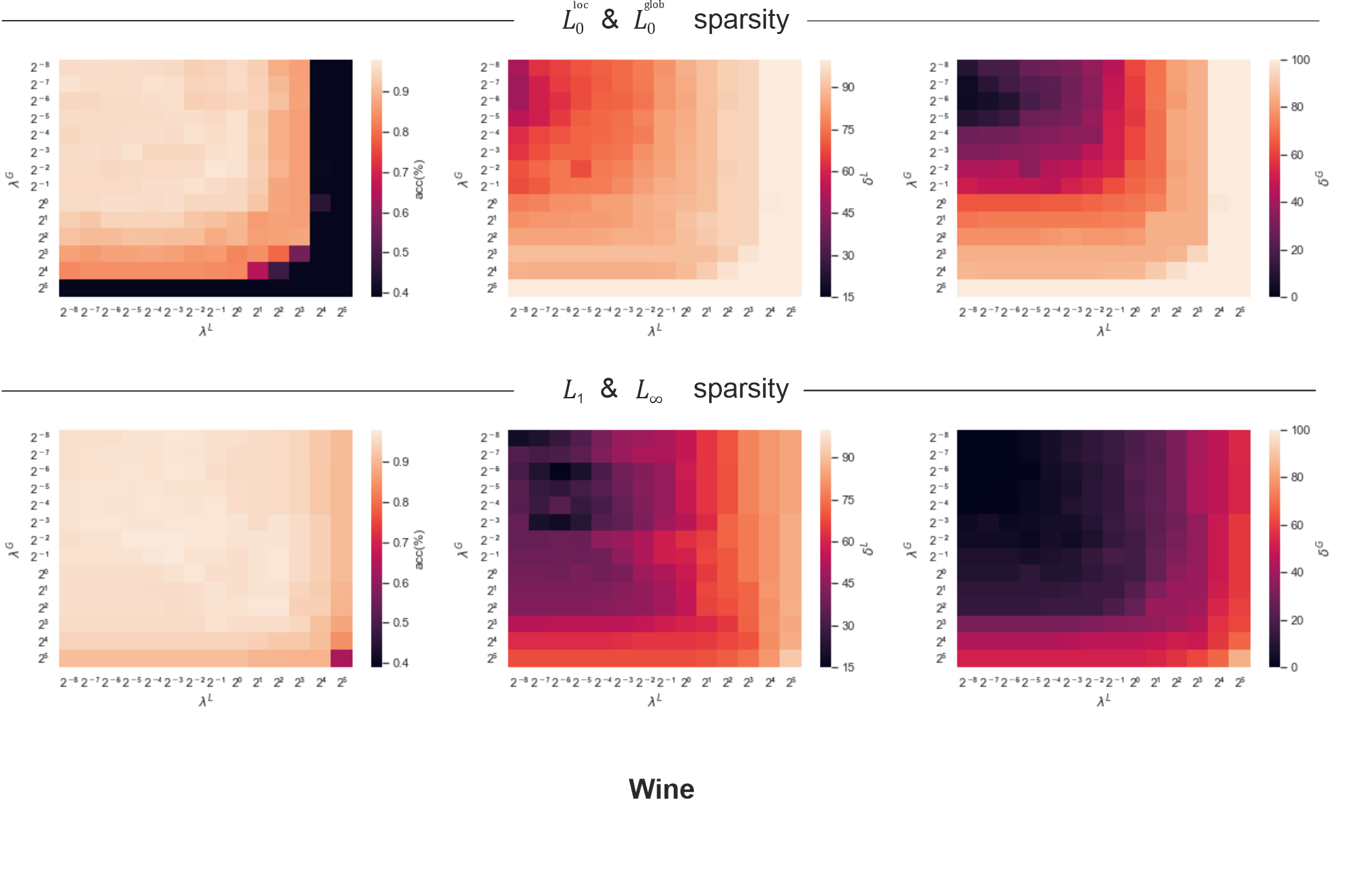}

\end{figure}


\end{document}